\documentclass[preprint,sort,compress,12pt]{elsarticle}

\usepackage{amssymb}
\usepackage{amsthm}
\usepackage{amsmath}
\usepackage{mathtools}
\usepackage{mathrsfs}
\usepackage{algorithm}
\usepackage{algpseudocode}
\usepackage{array}
\usepackage{multirow}
\usepackage{listings}
\usepackage{tabu}
\usepackage{enumerate}
\usepackage{lineno}
\usepackage{fullpage}
\usepackage{xcolor}
\usepackage[colorinlistoftodos]{todonotes}
\usepackage{bbm}
\usepackage[colorlinks=true]{hyperref}
\usepackage{url}
\usepackage{textcomp}
\usepackage{gensymb}
\usepackage{soul}
\usepackage{booktabs}
\usepackage{multirow}
\usepackage{arydshln}
\usepackage{graphicx}
\usepackage{subfig}
\usepackage{float}
\usepackage{caption}
\usepackage{tikz}
\usetikzlibrary{shapes}

\theoremstyle{definition}

\theoremstyle{remark}

\biboptions{numbers,comma,round,square}
\graphicspath{ {./figs/} }

\journal{Elsevier}

\begin{document}
\begin{frontmatter}

 \title{HUG-VAS: A Hierarchical NURBS-Based Generative Model for Aortic Geometry Synthesis and Controllable Editing}



\author[ndAME]{Pan Du}
\author[cornell]{Mingqi Xu}
\author[ndACMS]{Xiaozhi Zhu}
\author[ndAME,cornell]{Jian-xun Wang\corref{corxh}}

\address[ndAME]{Department of Aerospace and Mechanical Engineering, University of Notre Dame, Notre Dame, IN}
\address[cornell]{Sibley School of Mechanical and Aerospace Engineering, Cornell University, Ithaca, NY}
\address[ndACMS]{Department of Applied and Computational Mathematics and Statistics, University of Notre Dame, Notre Dame, IN}

\cortext[corxh]{Corresponding author. Tel: +1 540 3156512}
\ead{jw2837@cornell.edu}

\begin{abstract}

Accurate, patient-specific vascular geometry is pivotal for diagnosis, planning, and device design, yet existing statistical shape modeling (SSM) pipelines rely on linear priors and topology-specific preprocessing that limit realism, scalability, and interoperability. We present HUG-VAS, a Hierarchical NURBS Generative framework for Vascular models, that unifies NURBS-based 3D shape encoding with diffusion-based generative modeling to synthesize fine-grained, CFD-ready aortic anatomies. HUG-VAS factorizes shape into (i) vessel centerlines generated by a denoising diffusion model and (ii) cross-sectional radius profiles synthesized by a classifier-free guided diffusion model conditioned on the centerline, thereby decoupling and preserving stochastic variability across these two anatomical layers. Beyond unconditional synthesis, we enable training-free, zero-shot conditional generation via diffusion posterior sampling from image-derived prompts (e.g., sparse 3D points, slice contours, or partial surface patches), supporting interactive semi-automatic segmentation, editing and robust reconstruction under degraded imaging. Trained on 21 patient-specific MRA cases, HUG-VAS generates multi-branch aortas with supra-aortic vessels whose biomarker distributions closely match the source cohort, and whose watertight NURBS outputs directly integrate with downstream CFD solvers. To our knowledge, this is the first SSM framework that bridges image-derived priors and generative shape synthesis through a unified combination of NURBS parameterization, hierarchical diffusion, and DPS, enabling a practical path from limited clinical anatomic information to simulation-ready vascular geometry.

\end{abstract}

\begin{keyword}
Statistical Shape Modeling \sep NURBS \sep Generative AI \sep Semi-automatic Segmentation
\end{keyword}
\end{frontmatter}


\section{Introduction}
\label{sec:intro}
Cardiovascular diseases (CVDs) are the leading cause of death worldwide and span diverse pathologies, including aortic disorders, congenital heart diseases, and cerebrovascular events~\cite{nabel2003cardiovascular}. Accurate patient-specific characterization of vascular geometry and blood flow is critical for linking structural variability to pathological hemodynamics, thereby informing diagnosis, prognosis, and treatment planning~\cite{sel2024building}. Advances in medical imaging, such as magnetic resonance imaging (MRI) and computed tomography (CT), now enable high-resolution vascular reconstructions, supporting quantitative assessment of both anatomical and functional features. These anatomical representations provide essential inputs for downstream applications such as patient-specific computational fluid dynamics (CFD) simulations, which enable digital twins of individualized hemodynamics for precision diagnosis, therapy planning, and surgical strategy development~\cite{taylor2009patient,coorey2022health}. Shape-derived features further allow the prediction of biomarkers (e.g., wall shear stress, pressure gradients) via statistical or machine-learning models, aiding risk stratification and disease classification~\cite{marcos2021applications}. Moreover, synthetic cohorts generated from patient-specific geometries can accelerate image-based CFD through surrogate models and support medical device design~\cite{bridio2023generation}. Finally, 3D-printed vascular phantoms derived from these digital geometries provide a platform for bench-top flow experiments that inform surgical planning, device testing, and cardiovascular research~\cite{vukicevic2017cardiovascular}.

Despite its clinical importance, robust characterization of vascular anatomy remain challenging due to complex, multi-branch topology. Early shape analyses primarily focused on two-dimensional (2D) cross-sectional images, using size measurements, low-dimensional descriptors, or morphometric models of contours and anatomical lines~\cite{durrleman2009statistical, twining2005unified}. These methods provided insight into population-level variation but lacked full three-dimensional (3D) surface representations, limiting their utility for CFD or procedural planning. 
In parallel, algorithmic pipelines were developed to extract 3D geometries directly from medical images, using techniques such as level sets~\cite{leventon2000level}, marching cubes~\cite{masala2013improved}, and non-uniform rational B-splines (NURBS) based reconstructions~\cite{zhang2007patient}. These methods eventually matured into widely-used platforms such as SimVascular~\cite{updegrove2017simvascular} and 3D Slicer~\cite{kikinis20133d}. However, these workflows remain labor-intensive, typically producing a single geometry per patient per session, and are highly sensitive to imaging artifacts such as low resolution, incomplete coverage, or noise. Outcomes also vary across operators and repeated segmentations of the same case, reducing reliability and hindering routine adoption in diagnostic pipelines~\cite{valen2018real}.

Moving from representing individual shapes to modeling population variability, \emph{statistical shape modeling} (SSM) extends traditional shape analysis to full 3D cohorts. SSM establishes point-to-point correspondence across geometries and applies dimensionality reduction, most commonly principal component analysis (PCA), to extract dominant modes of variation. This enables morphometric analysis and synthesis of statistically consistent shapes, complementing image-based reconstruction with a modeling framework for anatomical diversity. Establishing correspondence, a prerequisite for PCA, requires encoding each sample in a consistent data structure (typically a fixed-length feature vector). The same framework has also been applied beyond geometry to images~\cite{deo2024few, dirix2024synthesizing}, inlet boundary conditions~\cite{saitta2023data, garzia2023coupling}, and biological signals~\cite{doste2022training, Myers2023MLsurrogateQSP}.
Numerical shape representations with SSM generally fall into two paradigms: \emph{deformation-based}~\cite{du2022deep, bruse2016non, bruse2017detecting, scarpolini2023enabling, cosentino2020statistical, sophocleous2022feasibility, Geronzi2023ShapeFeatures, durrleman2009statistical, niederer2020creation, mansi2011statistical, rodero2021linking, Verstraeten2024SyntheticAVStenosis, oguz2016entropy, Grassi2011femurMorphing, Li2024ImageUncertaintyCFD} and \emph{parameterization-based}~\cite{thamsen2021synthetic, alvarez2017tracking, Ostendorf2024SyntheticAorticDissection, romero2021clinically, Romero2025RobustVesselShape, liang2017machine, Wiputra2023thoracicAortaSSM, Maquart2021BREPmeshing, young2009computational, khalafvand2018assessment, metz2012regression, goubergrits2022ct, Verhulsdonk2024ShapeOfMyHeart, Varela2017LAanalysis, bridio2023generation, Bisighini2025stent, hoeijmakers2020combining, gambaruto2012decomposition, keustermans2018high, bruning2020characterization, Gahima2023UnfittedElasticBed, Ballarin2016PODGalerkinCFR}. Deformation-based approaches stem from shape registration techniques in computer vision~\cite{vaillant2005surface, lamecker2008variational}, where a template is diffeomorphically deformed to match targets, and the resulting vector fields (``currents'') compactly encode variation. They have been applied to the aorta~\cite{du2022deep, bruse2016non, bruse2017detecting, scarpolini2023enabling, cosentino2020statistical, sophocleous2022feasibility, Geronzi2023ShapeFeatures}, iliac arteries~\cite{Li2024ImageUncertaintyCFD}, aortic valves~\cite{Verstraeten2024SyntheticAVStenosis}, cardiac chambers~\cite{durrleman2009statistical, niederer2020creation, mansi2011statistical, rodero2021linking}, brain~\cite{oguz2016entropy}, and femur~\cite{Grassi2011femurMorphing}. Yet they require careful kernel tuning and often struggle to preserve watertight surfaces for complex branching; most applications remain single-channel. Limited multi-branch efforts (e.g., short supra-aortic branches with manual outlet landmarks~\cite{scarpolini2023enabling}; bifurcating aorto-iliac with centerline landmarks~\cite{Li2024ImageUncertaintyCFD}) highlight the difficulty of capturing full multi-branch morphometry.
Parameterization-based methods encode target anatomy with explicit mathematical constructs. For vascular systems, parameterizations are typically \emph{centerline-based} or \emph{surface-based}. Centerline-based models track centerline trajectories and assign cross-sectional profiles, ranging from single radii~\cite{thamsen2021synthetic} to elliptical~\cite{alvarez2017tracking, Ostendorf2024SyntheticAorticDissection} or spline curves~\cite{romero2021clinically, Romero2025RobustVesselShape}, then reconstruct the surface by interpolation; multi-branch coverage remains limited~\cite{thamsen2021synthetic, scarpolini2023enabling}. Surface-based models directly parameterize vessel walls, e.g., surface unwrapping for a single-channel aorta~\cite{liang2017machine}, extensions to simple branching~\cite{Wiputra2023thoracicAortaSSM}, or boundary-representation (B-Rep) meshing for bifurcations~\cite{Maquart2021BREPmeshing}. These approaches are often topology-specific, require heavy preprocessing, and can suffer from mesh-quality issues.

Most existing SSM frameworks, whether deformation- or parameterization-based, rely on PCA for dimensionality reduction. While effective for capturing dominant linear correlations, PCA constrains shape variation to a low-dimensional linear subspace. New geometries synthesized by sampling this subspace often lack diversity and fail to capture the nonlinear, multimodal character of true anatomical variability, which arises from complex biological, developmental, and pathological factors. To enforce plausibility, PCA-based models typically assume a predefined distribution for the coefficients, most often multivariate Gaussian. Alternative sampling strategies, such as bootstrap, uniform, or generative adversarial networks (GANs)-based sampling, have been explored \cite{romero2021clinically}. Uniform sampling broadens variability, while Gaussian and GAN sampling yield more realistic shapes by accounting for coefficient covariances. Nonetheless, all such approaches remain limited by the linear PCA space and cannot directly learn or represent the underlying nonlinear distribution of anatomical shapes.

Generative models, grounded in probabilistic learning, have emerged as powerful tools for capturing complex data distributions. They have achieved notable success in image and video generation, large language models, and other CV tasks~\cite{liu2024survey, yazdani2023survey}, yet their application to anatomical SSM remains relatively limited. Unlike PCA-based approaches with predefined priors, generative models directly learn the underlying distribution, offering a principled alternative for synthesizing realistic and diverse shapes. Initial efforts in anatomy-focused SSM have centered on GANs~\cite{Landoll2024VirtualPatientECMO, Danu2019SyntheticVesselSurfaces, Wolterink2018BloodVesselGAN} and variational autoencoders (VAEs)~\cite{Dou2025VirtualChimera, kalaie2023graphLV, Kalaie2025RefinableShapeGeneration, Beetz2022MultiDomainVAE, Beetz2022InterpretMeshVAE, Feldman2023VesselVAE, Feldman2025VesselGPT}. These approaches typically require preprocessing steps to encode geometries into unified representations suitable for training. GANs learn by adversarially matching a generator against a discriminator to produce samples indistinguishable from training data. For example, Wolterink et al.~\cite{Wolterink2018BloodVesselGAN} parameterized coronary vessels as centerline coordinate–radius sequences and trained a GAN to synthesize plausible single-channel geometries. Danu et al.~\cite{Danu2019SyntheticVesselSurfaces} applied both GANs and VAEs to generate short single-channel vessels from 3D images or centerline–radius inputs. While GANs can produce visually realistic shapes, they scale poorly to complex vascular topologies and are prone to training instability and mode collapse, reducing both diversity and anatomical fidelity.
VAE-based SSM have primarily focused on cardiac chambers~\cite{Dou2025VirtualChimera, kalaie2023graphLV, Kalaie2025RefinableShapeGeneration, Beetz2022MultiDomainVAE, Beetz2022InterpretMeshVAE}, brain vasculature~\cite{Feldman2023VesselVAE}, and aortic geometries \cite{Feldman2025VesselGPT}. A standard VAE employs an encoder–decoder architecture with a Gaussian latent space; new shapes are sampled from this distribution and decoded back into geometry. Variants include: (i) a $\beta$-VAE with shape matching for left ventricle synthesis~\cite{kalaie2023graphLV, Kalaie2025RefinableShapeGeneration}; (ii) a convolutional mesh autoencoder (CoMA) within a multi-channel VAE for full heart chambers~\cite{Dou2025VirtualChimera}; and (iii) an unstructured VAE integrating surface parameterization with probabilistic encoding~\cite{Beetz2022MultiDomainVAE, Beetz2022InterpretMeshVAE}. For vessels, Feldman et al.~\cite{Feldman2023VesselVAE} proposed a recursive VAE to sequentially generate centerline nodes and contours, while their follow-up work~\cite{Feldman2025VesselGPT} used a GPT2-inspired sequential generator with radial splines to model aortic aneurysms. Despite these advances, VAE-based methods often emphasize branching topology over surface fidelity. This is problematic for large vessels such as the ascending aorta and arch, where precise local morphology is critical. Moreover, VAEs are prone to oversmoothing and limited latent expressiveness, leading to blurry reconstructions and difficulty capturing complex or multimodal variation.

Diffusion models have emerged as a leading class of generative methods, repeatedly demonstrating superior performance to VAEs and GANs across many domains~\cite{dhariwal2021diffusion}. Inspired by nonequilibrium thermodynamics, diffusion models synthesize data by reversing a forward process that incrementally corrupts samples with Gaussian noise. This iterative denoising is comparatively stable to train, tolerant to irregular data, and naturally supports conditional generation via auxiliary inputs or prompts \cite{chen2024overview}. While diffusion models have transformed CV and more recently scientific modeling~\cite{du2024confild,fan2025neural,liu2025confild}, its use in anatomical SSM remains in its infancy. Early work targets the heart \cite{Kadry2024DiffusionDigitalTwins} and several vascular systems, including cerebral, capillary, and retinal vessels \cite{Sinha2024TrIND, Deo2024FewShotCerebralAneurysm, Kuipers2024ConditionalSetDiffusion, Prabhakar2024VesselGraphDD}. Kadry et al.~\cite{Kadry2024DiffusionDigitalTwins} demonstrate conditional latent diffusion for 2D cardiac images, enabling controlled variation of scale and regional anatomy. Most vascular-focused diffusion studies emphasize connectivity and branching over high-fidelity surfaces. For example, Sinha et al.~\cite{Sinha2024TrIND} encode anatomy as signed distance fields (SDFs) with implicit neural representations (INRs) and diffuse in INR parameter space; Deo et al.~\cite{Deo2024FewShotCerebralAneurysm} generate cerebral aneurysms from binary/SDF image inputs; Kuipers et al.~\cite{Kuipers2024ConditionalSetDiffusion} diffuse sequences of centerline nodes with radii to grow hierarchical trees; and Prabhakar et al.~\cite{Prabhakar2024VesselGraphDD} denoise vessel graphs via separate node/edge steps. These formulations capture tree structure but under-specify large-vessel surface details. Addressing this gap requires domain-specific diffusion frameworks capable of capturing both topological and morphological realism, beyond the representational limits of INR or graph abstractions.

As highlighted earlier, SSM always operates within a chosen representation space, and the quality of generated results depends as much on the representation as on the learning algorithm. Existing studies have adopted diverse encodings, INR, explicit centerline-based profiles, or unstructured surface meshes, each optimized for specific tasks. This heterogeneity leads to fragmentation and poor interoperability, while most pipelines ultimately yield unstructured meshes that are memory-intensive, difficult to edit, and not readily reusable. To overcome these limitations, we advocate \emph{non-uniform rational B-splines} (NURBS) as a unifying backbone \cite{zhang2007patient}. NURBS provide smooth, watertight parameterizations through compact sets of control points and weights, supported by mature CAD tooling for editing, meshing, and coupling to isogeometric analysis. These properties make NURBS both simulation-ready and ML–friendly, enabling efficient integration with generative models while preserving post-edit flexibility not available with voxel or unparameterized mesh outputs.

Another limitation of existing vascular SSM methods is the treatment of vessel centerlines and radius profiles as a deterministic pair. In reality, local radius is influenced not only by geometric layout but also by hemodynamic and tissue constraints, so a single centerline can correspond to multiple plausible radial distributions. Existing sequential generators that output centerline–radius jointly \cite{Feldman2025VesselGPT} risk collapsing this variability. To better capture inter-patient diversity, we propose a \emph{hierarchical generative formulation} in which centerlines and radii are modeled by separate diffusion processes, with the latter conditioned on the former. This design preserves probabilistic variability in radial profiles for a fixed centerline, improving realism and enabling uncertainty-aware morphometric analysis.

Beyond model architecture, an orthogonal yet underexplored dimension is integrating SSM with imaging. Kadry et al.~\cite{Kadry2024DiffusionDigitalTwins} demonstrated conditional diffusion for anatomical image editing, suggesting a paradigm where vascular geometries are generated under user-defined constraints. This motivates the idea of \emph{semi-automatic segmentation}, where sparse cues from medical images (e.g., points, contours, or patches) guide a generative model to produce realistic, watertight geometries for a specific patient. Such capability is urgently needed: manual segmentation, though accurate, is slow and labor-intensive, whereas automated ML/DL methods are faster but often not robust, suffering from artifacts, poor generalization, and reliance on imperfect training labels~\cite{wolterink2016dilated, xia2019automatic,du2025ai,an2025hierarchical, wickramasinghe2020voxel2mesh,zhao2022segmentation, Kong2021WholeHeartMesh,sveinsson2025seqseg}. Their outputs, typically voxel grids or unparameterized meshes, frequently contain discontinuities and surface defects, requiring substantial manual repair before CFD use. What is lacking is a conditional generation framework that can translate sparse, noisy image-derived cues into high-quality, parameterized geometries that are watertight, simulation-ready, and easily editable. Closing this gap would not only streamline segmentation but also support many-query tasks such as inverse design, uncertainty quantification, and the synthesis of diverse training data for surrogate modeling. Recent work such as AortaDiff~\cite{an2025aortadiff} has taken steps in this direction, applying conditional diffusion for direct aortic surface reconstruction from CT/MRI volumes and producing CFD-compatible meshes with limited training data. However, its emphasis on automated reconstruction leaves open challenges in flexible representation, user-guided editing, and probabilistic modeling of centerline–radius variability.

\begin{figure}[htb!]
    \centering
    \includegraphics[width=0.99\textwidth]{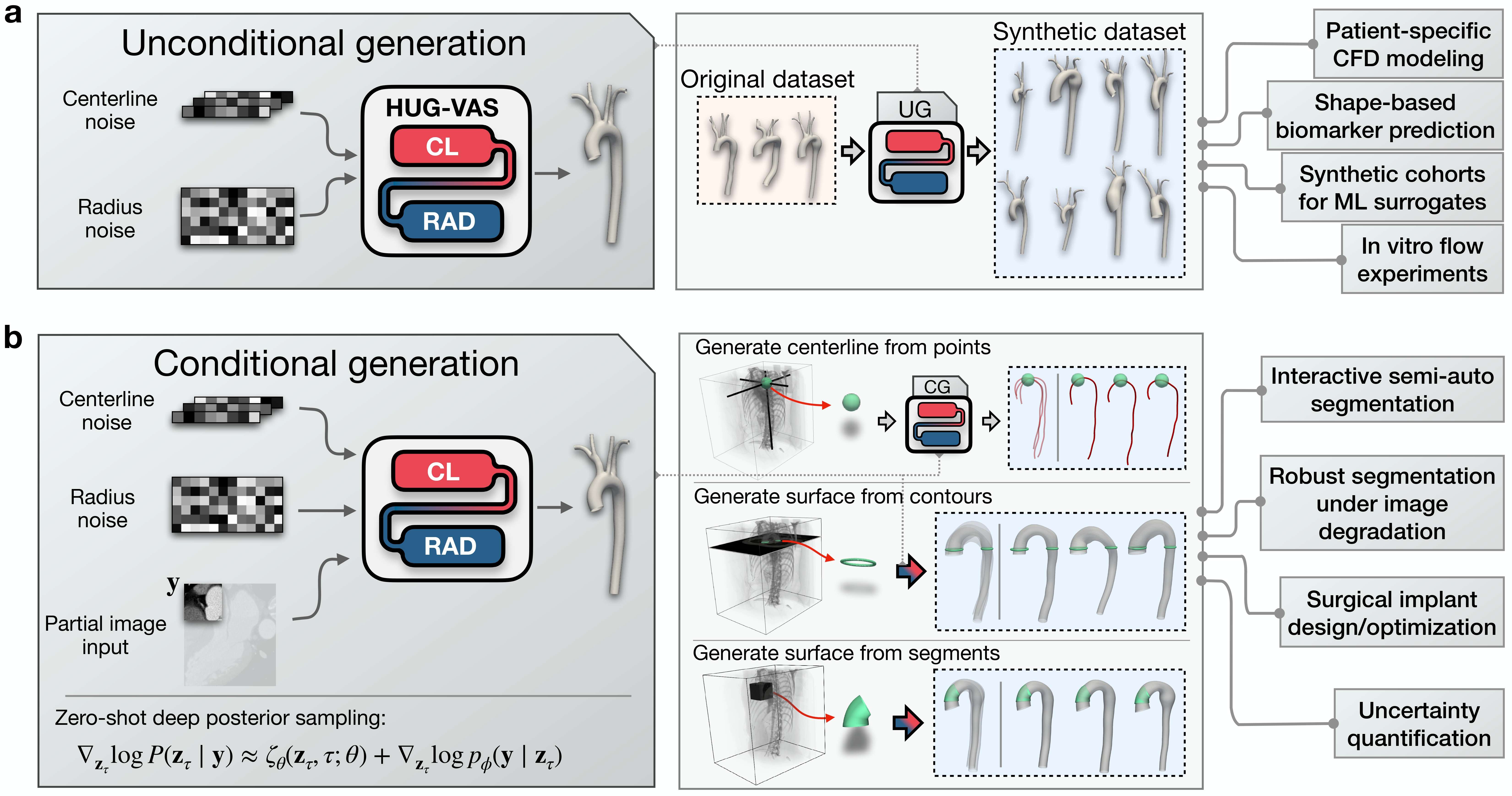}
    \caption{\textbf{a}, Unconditional generation: HUG-VAS takes as input random noise vectors for both centerline and radial encodings and synthesizes multi-branch aortic geometries that resemble anatomically plausible patient-specific shapes. The generated dataset expands the original cohort and enables a variety of downstream applications, including patient-specific CFD modeling, shape-based biomarker prediction, synthetic cohort generation for machine learning surrogates, and in vitro flow experiments. \textbf{b}, Conditional generation: HUG-VAS incorporates partial image observations and performs zero-shot posterior sampling to synthesize anatomically consistent aortic geometries that satisfy the given condition. It supports conditional generation of centerlines from sparse 3D points, and surface reconstruction from contours or surface patches. These capabilities enable applications such as interactive semi-automatic segmentation, robust segmentation under image degradation, surgical implant design and optimization, and uncertainty quantification. Together, these two modes of generation make HUG-VAS a versatile framework that serves both as a traditional statistical shape model and as an intelligent geometry constructor that bridges to image-derived priors.}
    \label{fig:usage}
\end{figure}
In response to the challenges outlined above, we introduce the \emph{Hierarchical NURBS Generative framework for Vascular models} (HUG-VAS), a high-fidelity generative approach for vascular geometry synthesis that combines enhanced variability, robustness, and conditional generation capabilities (Fig.~\ref{fig:usage}). HUG-VAS integrates NURBS with diffusion modeling in a hierarchical manner, enabling probabilistic learning within a compact and interpretable parameter space. NURBS parameterization ensures anatomically accurate, watertight surfaces that are inherently editable and simulation-ready, while their implementation within an automatic differentiation (AD) framework facilitates integration into optimization workflows such as hemodynamic tuning or surgical implant design. Unlike prior deep learning–based generative models, HUG-VAS adopts a hierarchical architecture inspired by Zeng et al.~\cite{zeng2022lion}. A denoising diffusion probabilistic model (DDPM) generates vessel centerlines (CL), and a classifier-free guided diffusion model generates radius (RAD) profiles conditioned on the centerline. This hierarchical setup preserves stochastic variability in radii for a fixed centerline, thereby improving both anatomical diversity and geometric realism. In addition, by leveraging Bayesian diffusion posterior sampling, HUG-VAS supports zero-shot conditional generation, enabling semantically consistent geometries under diverse user-defined constraints without retraining. This flexibility unlocks a wide range of applications, from semi-automatic segmentation and robust image-based reconstruction to surgical design and uncertainty quantification. In this work, we demonstrate both unconditional and conditional generation of the thoracic aorta with supra-aortic branches, including the Left Subclavian Artery (LSA), Left Common Carotid Artery (LCCA), Right Subclavian Artery (RSA), and Right Common Carotid Artery (RCCA), as detailed in Section~\ref{sec:result}. To the best of our knowledge, HUG-VAS represent the first attempt of SSM that bridges image-derived priors with generative shape synthesis while unifying a standard NURBS representation with a hierarchical diffusion architecture.

The remainder of the paper is organized as follows: Section 2 presents an overview of the HUG-VAS architecture and showcases its unconditional and conditional generation results. Section 3 evaluates the generation quality, discusses current limitations, and outlines future directions. Section 4 provides a detailed description of the methodology.

\section{Results}
\label{sec:result}
\subsection{HUG-VAS framework and case setup} 
\label{sec:hugvas}

The HUG-VAS framework comprises two main components: a NURBS-based parameterization module for vascular encoding and a hierarchical latent diffusion module for generative modeling. We represent the aorta with supra-aortic branches as a vascular graph $\mathbf{\Psi} = \{\mathbf{V}, \mathbf{E}\}$, where $\mathbf{V} = \{\mathbf{v}^i\}_{i=1}^b$ denotes a set of single-channel vessel surfaces, and $\mathbf{E}$ stores their connectivity. Since the branching topology is fixed, we define $\mathbf{E} = \{e_i\}_{i=1}^f$, where each scalar $e_i$ specifies the relative bifurcation location along the parent vessel. For the thoracic aorta, $b=5$ and $f=4$ correspond to the main aorta and four supra-aortic branches (LSA, LCCA, RSA, RCCA) with their bifurcations. Each vessel $\mathbf{v}^i$ is parameterized by a set of NURBS-based centerline control points $\mathbf{C}^i \in \mathbb{R}^{n_i \times 3}$ and corresponding cross-sectional radii $\mathbf{R}^i \in \mathbb{R}^{n_i \times m_i}$, where $n_i$ and $m_i$ are the streamwise and angular discretizations, respectively. This parameterization involves a sequence of algorithmic steps including centerline extraction, B-spline curve fitting, discretional frame stratification, skeleton estimation, and NURBS surface fitting, as detailed in Section~\ref{sec:meth}. In general, The encoding maps a vascular sample $\mathbf{\Psi}$ into a latent representation $\mathbf{z}$:
\begin{equation}
    \mathbf{z} = \mathcal{E}\left(\mathbf{\mathbf{\Psi}}\right) = \{ \{\mathbf{C}^i, \mathbf{R}^i\}_{i=1}^b, \mathbf{E} \}.
\end{equation}
Given a training dataset $\mathcal{D} = \{\mathbf{\Psi}^j\}_{j=1}^N$ of size $N$, we train a hierarchical diffusion model for each vessel branch $i$ using $\{\mathbf{C}^{i,j}, \mathbf{R}^{i,j}\}_{j=1}^N$. For each branch-level model, a standard DDPM learns the distribution of centerline control points $\{\mathbf{C}^{i,j}\}_{j=1}^N$, and a classifier-free guided diffusion model learns the distribution of radial profiles $\{\mathbf{R}^{i,j}\}_{j=1}^N$, conditioned on the centerline $\mathbf{C}^i$. After training, new centerlines and radii are generated in two stages:
\[
\begin{cases}
    \hat{\mathbf{C}} = \eta^c_{\tau=T} \cdots \eta^c_{\tau=2} \circ \eta^c_{\tau=1} \left( \mathbf{C}_0 \right) \\
    \hat{\mathbf{R}} = \eta^r_{\tau=T} \left(\hat{\mathbf{C}},  \cdots \eta^r_{\tau=2} \left(\hat{\mathbf{C}},  \eta^r_{\tau=1} \left(\hat{\mathbf{C}} , \mathbf{R}_0 \right)\right)\right), 
\end{cases}
\]
where $\eta^c$ and $\eta^r$ are the trained denoising neural networks for centerline and radius, respectively, and $T=1000$ is the diffusion horizon. The generated $\hat{\mathbf{C}}$ and $\hat{\mathbf{R}}$ are used to reconstruct smooth vessel surfaces via NURBS.
The full vascular model is then assembled by sampling bifurcation statistics $\{\mathbf{E}^j\}_{j=1}^N$and merging the reconstructed vessels with boolean operations. This decoding pipeline is defined as:
\begin{equation}
    \mathbf{\Psi} = \mathcal{D}\left(\mathbf{\mathbf{z}}\right) = \mathcal{A}\left( \left\{\hat{\mathbf{v}}^i\right\}_{i=1}^b, \hat{\mathbf{E}} \right), = \mathcal{A}\left( \left\{\mathcal{B}\left(\hat{\mathbf{C}}^{i}, \hat{\mathbf{R}}^{i}\right)\right\}_{i=1}^b, \hat{\mathbf{E}} \right),
\end{equation}
where $\mathcal{B}$ denotes NURBS surface construction, and $\mathcal{A}$ denotes vessel assembly process. 

HUG-VAS also features zero-shot conditional generation, where sample synthesis can follow user-specified prompts $\mathbb{\mathbf{y}}$ without retraining. This is formulated as Bayesian inference, with the diffusion model providing the prior $p(\mathbf{z})$ and the conditioned samples are drawn from the posterior $p(\mathbf{z}\mid \mathbf{y}) \propto p(\mathbf{y}\mid \mathbf{z}) p(\mathbf{z})$. This capability relies on a differentiable forward map $\mathcal{F}$ that relates a full-state sample $\mathbf{z}$ to its corresponding prompt. The map is defined as a composition of two steps: decoding $\mathbf{z}$ into a vessel surface $\mathbf{v}$, and projecting $\mathbf{v}$ to the prompt space via an observation function $\mathcal{O}$, i.e., $\mathbf{y} = \mathcal{F}(\mathbf{z}) = \mathcal{O}(\mathcal{B}(\mathbf{z}))$. 
During denoising, HUG-VAS computes the gradient of $\mathcal{F}$ with respect to $\mathbf{z}$ via AD and uses it to iteratively adjust the score function, steering the sampling trajectory toward posterior-consistent samples $\hat{\mathbf{z}}$ that satisfy the prompt $\mathbf{y}$. This mechanism, known as Deep Posterior Sampling (DPS), sequentially updates centerline and radius variables in our hierarchical setup,
\[
\begin{cases}
\nabla_{\mathbf{C}_\tau} \log p(\mathbf{C}_\tau \mid \mathbf{y}_C) 
\approx \hat{\mathbf{\zeta}}^c_{\theta}(\mathbf{C}_\tau, \tau ; \theta) 
+ \nabla_{\mathbf{C}_\tau} \log p(\mathbf{y}_C \mid \mathbf{C}_\tau) \\
\nabla_{\mathbf{R}_\tau} \log p(\mathbf{R}_\tau \mid \mathbf{y}_R) 
\approx \hat{\mathbf{\zeta}}^r_{\theta}(\mathbf{R}_\tau, \hat{\mathbf{C}},  \tau ; \theta) 
+ \nabla_{\mathbf{R}_\tau} \log p(\mathbf{y}_R \mid \mathbf{R}_\tau),
\end{cases}
\]
where $\hat{\mathbf{\zeta}}^c_{\theta}$, $\hat{\mathbf{\zeta}}^r_{\theta}$ denote the learned score functions for centerlines and radii, and $\mathbf{y}_C$, $\mathbf{y}_R$ are the respective prompts. Note that DPS is applied sequentially to centerline and radius, but both stages can be batch-parallelized to generate multiple samples efficiently, similar to standard single-stage DPS. Further methodological details are provided in Section~\ref{sec:meth}.

\paragraph{Dataset} We evaluate HUG-VAS on the Vascular Model Repository (VMR)~\cite{Wilson2013}, an open-source collection of normal and diseased anatomies. We selected 21 human cases covering healthy aortas, thoracic aneurysms, and post-Fontan congenital heart disease. Since the original surface segmentations in the repository provided limited coverage of the supra-aortic branches, we manually re-segmented all geometries from the original CMR image data using SimVascular~\cite{updegrove2017simvascular}. Each case includes five vessels (aorta, LSA, LCCA, RSA, RCCA) with consistent topology.

\paragraph{Evaluation scenarios} Using the trained HUG-VAS model, we present unconditional and conditional generation results. Unconditional generation synthesizes diverse aortic anatomies from random noise, enabling cohort expansion for surrogate modeling and biomarker studies. Conditional generation supports a variety of image-driven tasks on out-of-training patients, including generating centerlines from sparse point prompts, reconstructing surfaces from 2D contours, completing geometries from partial meshes, recovering shapes from low-resolution scans, and surgical planning pipelines for thoracic aneurysms. These scenarios are designed to showcase the generative capability of HUG-VAS and its potential to bridge image-derived priors with SSM, supporting interactive segmentation, robust reconstruction under degraded imaging, quantification of anatomical uncertainty, and shape optimization. 

\subsection{Aortic shape synthesis through unconditional generation} 
We first evaluate the unconditional generation capability of HUG-VAS for synthesizing aortas with supra-aortic branches. This setting demonstrates the model's ability to learn the intrinsic anatomical distribution and to generate diverse yet plausible vascular geometries without external constraints. This is valuable for expanding training cohorts, enabling data augmentation for ML tasks, and exploring the natural variability of aortic anatomy.

\begin{figure}[t!]
    \centering
    \includegraphics[width=0.99\textwidth]{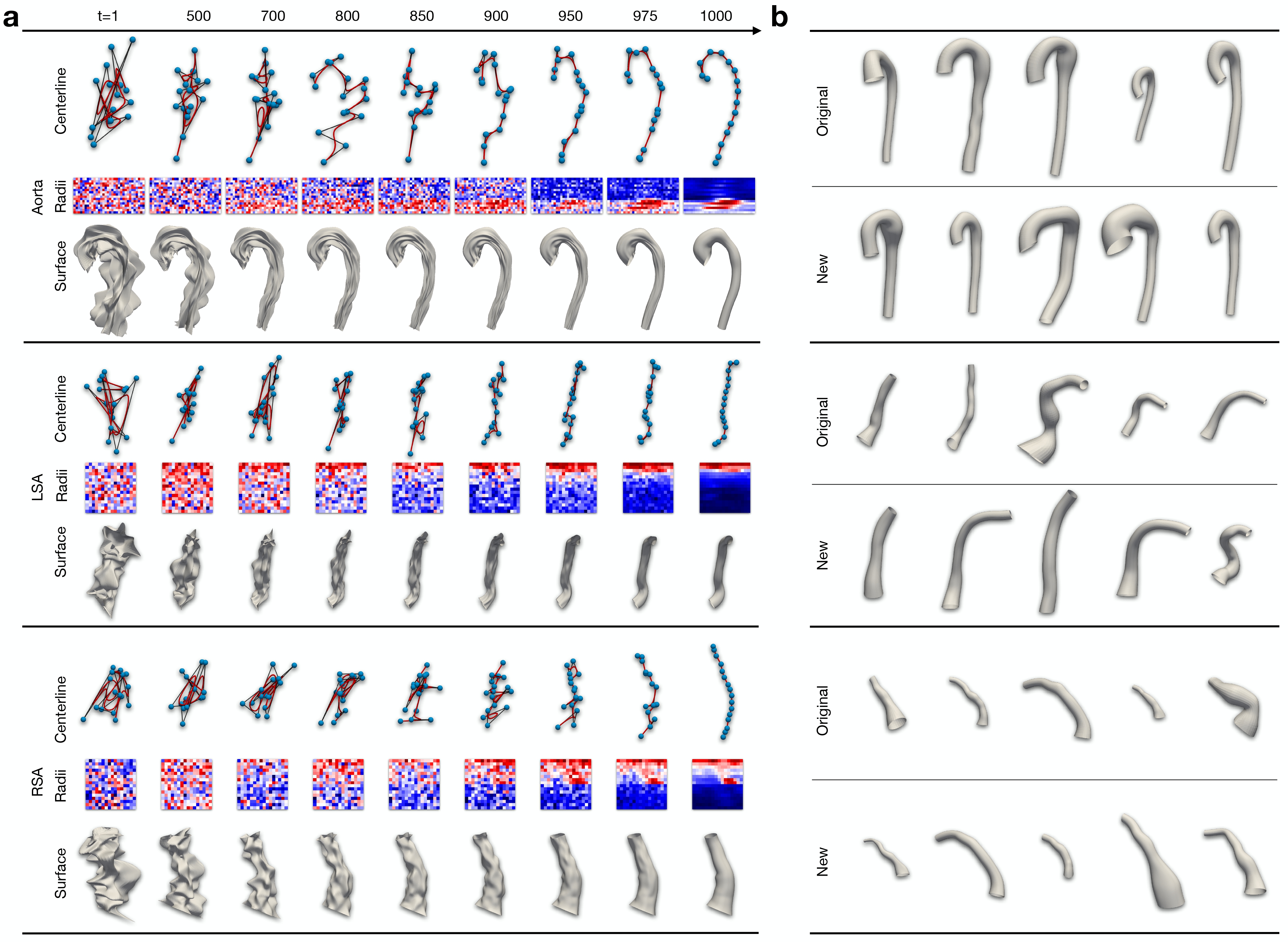}
    \caption{\textbf{a}, Visualization of the HUG-VAS denoising process across diffusion time steps (from $\tau=1$ to $\tau=1000$), showing progressive refinement of centerline control points (top row), radial profiles (middle row), and full surface reconstructions (bottom row). Results are shown for three representative branches: the main aorta (top panel), left subclavian artery (LSA, middle panel), and right subclavian artery (RSA, bottom panel). \textbf{b}, Comparison between original patient-specific geometries (top rows) and newly synthesized samples (bottom rows) for each anatomical region. Synthesized shapes exhibit strong anatomical plausibility and diversity, closely reflecting the morphology observed in the original dataset. Together, these results highlight the multi-stage generative capability of HUG-VAS and its ability to model anatomically coherent variations across both primary and branching vessels.}
    \label{fig:ug1}
\end{figure}

Figure~\ref{fig:ug1} illustrates the unconditional generation process with representative generated samples for the aorta, LSA, and RSA. For each vessel, the denoising trajectories of centerline control points, radial profiles, and reconstructed surfaces are visualized in consecutive rows from top to bottom, as illustrated in Fig.~\ref{fig:ug1}a. In the centerline plots, control points, polygons, and the resulting cubic B-spline curves are shown in blue, black, and red, respectively. 
Starting from noise, the control points gradually align into ordered sequences, converging to anatomically coherent structures such as the ascending–arch–descending configuration of the aorta. We use a resolution of 100 points to construct the smooth centerline curve, while the corresponding control point set contains only $n = 16$, yielding a $16\%$ compression ratio. Correspondingly, the radial profiles are encoded as $n \times m$ images, where $n=16$ cross-sectional locations are sampled uniformly along the centerline and $m=32$ radii are defined per section. The discretization parameters ($n$, $m$) used for the supra-aortic branches are summarized in Tab.~\ref{tab:latentdimension}. Each row of the image corresponds to one cross-section, with color intensity representing radial extent, and the vertical axis (from bottom to top) corresponds to the streamwise direction of the main aorta, i.e., from inlet to outlet. During denoising, the noisy profiles resolve into smooth tapering patterns, with higher values upstream (in red) and smaller radii downstream (in blue), consistent with physiological narrowing. Abnormalities such as aneurysms or coarctations manifest as localized deviations in this image, making the representation interpretable. For smaller branches such as the LSA and RSA, $m$ is reduced to $16$ due to fewer surface features, but the tapering behavior is similarly captured. Surface reconstructions are generated by combining the final centerlines with their radial profiles through NURBS fitting. The surfaces evolve from noisy, irregular shapes into globally smooth, anatomically realistic vessels. 

Figure~\ref{fig:ug1}b compares original patient-specific geometries with newly synthesized samples. The generated shapes exhibit strong anatomical plausibility and diversity, with variations in curvature, arch size, branch length, and tapering patterns. For example, synthesized aortas display both healthy and aneurysmal morphologies, while LSA and RSA samples vary in orientation and bending. This diversity arises from the hierarchical architecture of HUG-VAS, which decouples centerline and radial generation, enabling fine-grained variability in both structure and surface detail. Results for the remaining branches (RCCA and LCCA) are presented in Supplementary Note 2.

\begin{figure}[htp!]
    \centering
    \includegraphics[width=0.99\textwidth]{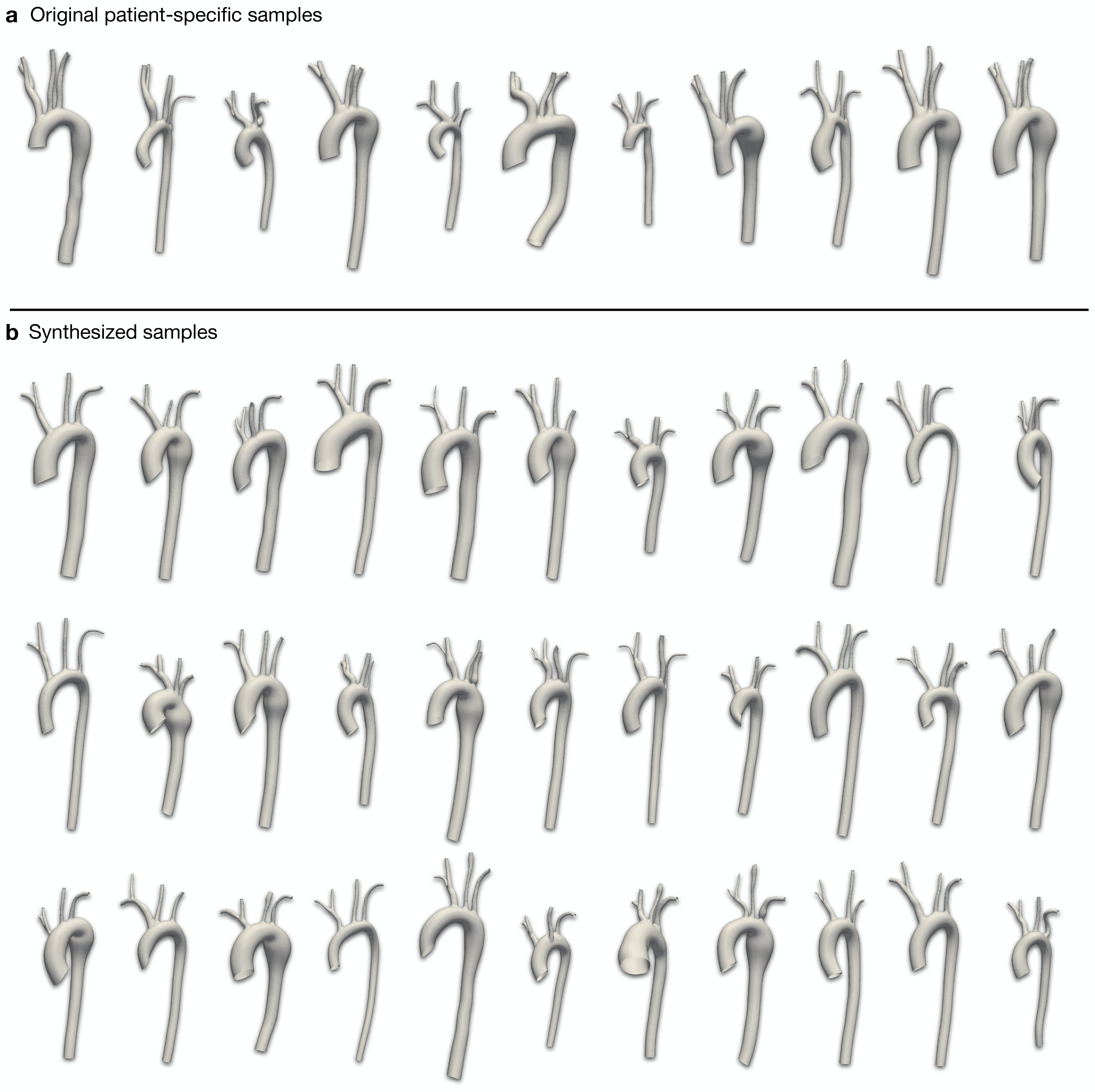}
    \caption{\textbf{a}, Gallery of representative original patient-specific aortic geometries, including anatomical variations across the main aorta and supra-aortic branches. \textbf{b}, Gallery of synthesized aortic geometries generated by HUG-VAS, exhibiting high anatomical plausibility, structural coherence, and diversity across samples.}
    \label{fig:ug2}
\end{figure}
After synthesizing individual vessels, we assemble complete multi-branch anatomies by sampling bifurcation statistics and merging the reconstructed surfaces. Figure~\ref{fig:ug2} presents galleries of original and generated aortas with supra-aortic branches. The results show a wide spectrum of morphologies, including differences in arch size, branch spacing, and bifurcation locations, while preserving correct topology and watertight surfaces. Moreover, thanks to our automated post-processing pipeline, the generated geometries are open-ended and boundary-aware (i.e., inlet and outlet patches are preserved), making them directly meshable for CFD simulations without further repair, an important convenience not offered by prior methods. An extended gallery of synthesized multi-branch anatomies is provided in Supplementary Note 3.
Together, these results demonstrate that HUG-VAS can unconditionally synthesize anatomically faithful, diverse, and CFD-ready vascular geometries, supporting both morphometric exploration and data augmentation for downstream learning tasks. 

\subsection{Conditional generation of centerline from point prompts} 
\begin{figure}[htb!]
    \centering
    \includegraphics[width=1.0\textwidth]{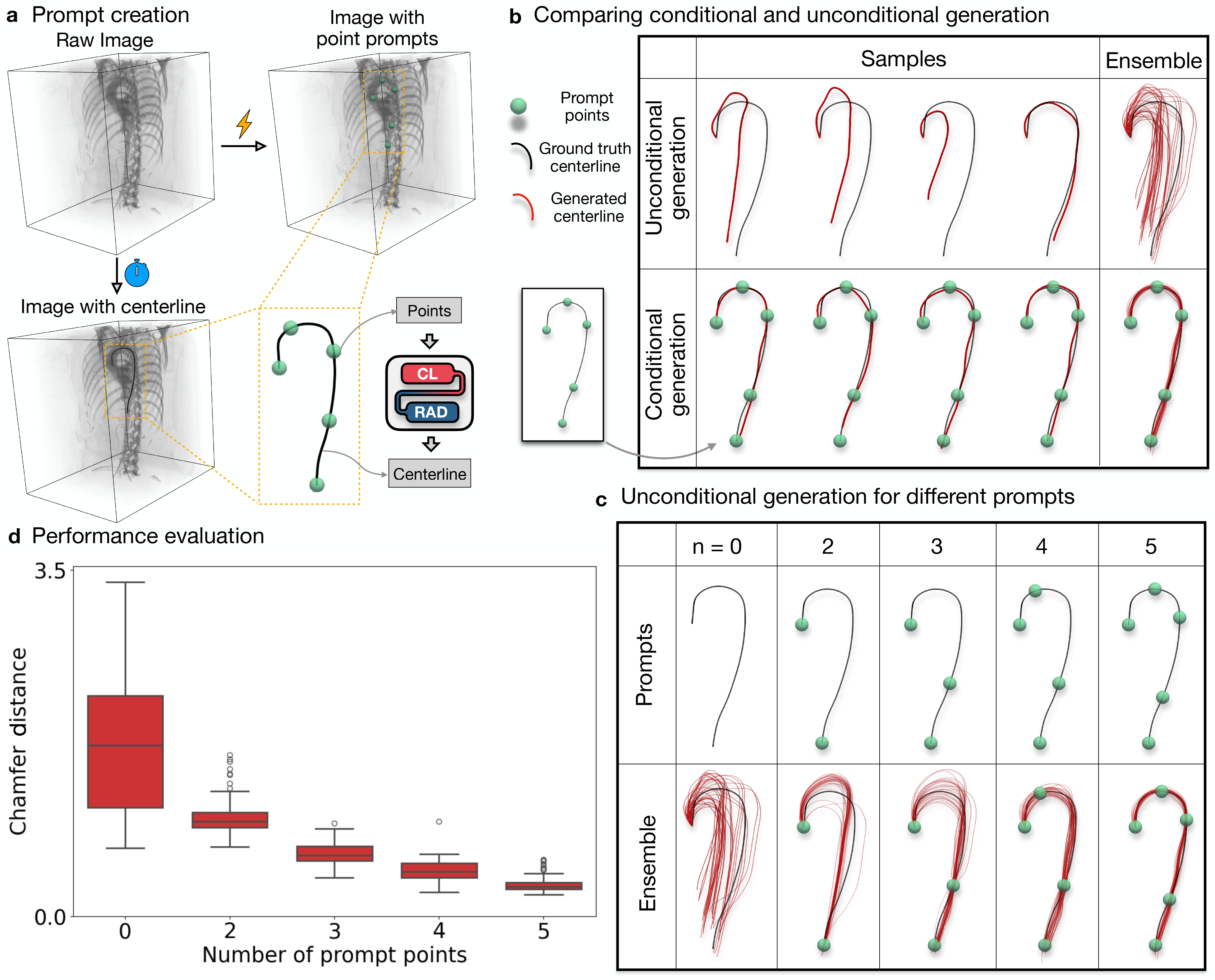}
    \caption{\textbf{a}, Schematic of HUG-VAS generating an aortic centerline from user-defined point prompts. Instead of manually segmenting the full centerline from the raw image (left path), the user selects a sparse set of points (green), which are passed into HUG-VAS (CL: centerline model, RAD: radial profile model) to synthesize the centerline. \textbf{b}, Comparison of unconditional (top) and conditional (bottom) generation. Conditional samples align with the prompts and closely follow the ground truth (black), whereas unconditional samples exhibit broader variability. \textbf{c}, Ensembles generated under different numbers of centerline point prompts ($n=0$–5). Adding more prompts progressively tightens the ensembles and improves alignment with the ground truth. \textbf{d}, Quantitative evaluation of generation accuracy. Chamfer distance between generated samples and the ground truth decreases monotonically as the number of prompts increases, reflecting both improved fidelity and reduced uncertainty. (All results are shown on test cases not seen during training)}
    \label{fig:cg1}
\end{figure}

HUG-VAS also features training-free conditional generation, in which user-defined prompts guide synthesis toward anatomies that satisfy given constraints. To demonstrate this capability, we first consider the task of generating aortic centerlines from sparse point prompts (Fig.~\ref{fig:cg1}). Traditionally, centerlines are often extracted manually from volumetric images, a process that is both time-consuming and labor-intensive. Here, our framework allows the user to provide only a small set of points, through which the model generates a complete centerline that conforms to both the prompts and learned anatomical priors. Note that all evaluations are performed on test cases not seen during training of the HUG-VAS model.

Figure~\ref{fig:cg1}a illustrates how point prompts are assimilated within the HUG-VAS pipeline. When no prompts are given, unconditional generation produces an ensemble of anatomically plausible but diverse centerlines that broadly cover the space around the true anatomy (Fig.~\ref{fig:cg1}b, top). Adding point constraints narrows this ensemble: the generated centerlines are guided to pass through the specified points and cluster closely around the ground-truth curve (Fig.~\ref{fig:cg1}b, bottom). This effect becomes more pronounced as the number of prompts increases, leading to ensembles that are progressively tighter and more faithful to the true anatomy (Fig.~\ref{fig:cg1}c). 
The improvement can be quantified using the Chamfer distance between generated and ground-truth centerlines (Fig.~\ref{fig:cg1}d). As the number of prompts increases from 0 to 5, the average Chamfer distance monotonically decreases from 1.71 to 0.33, while the standard deviation drops from 0.72 to 0.09, demonstrating both higher fidelity and reduced variance. Thus, conditional generation not only aligns the outputs with user-provided information but also reduces uncertainty, enabling the ensemble itself to serve as a robust segmentation result with quantified uncertainty.

\subsection{Conditional generation of surface from contour prompts} 

\begin{figure}[htb!]
    \centering
    \includegraphics[width=1.0\textwidth]{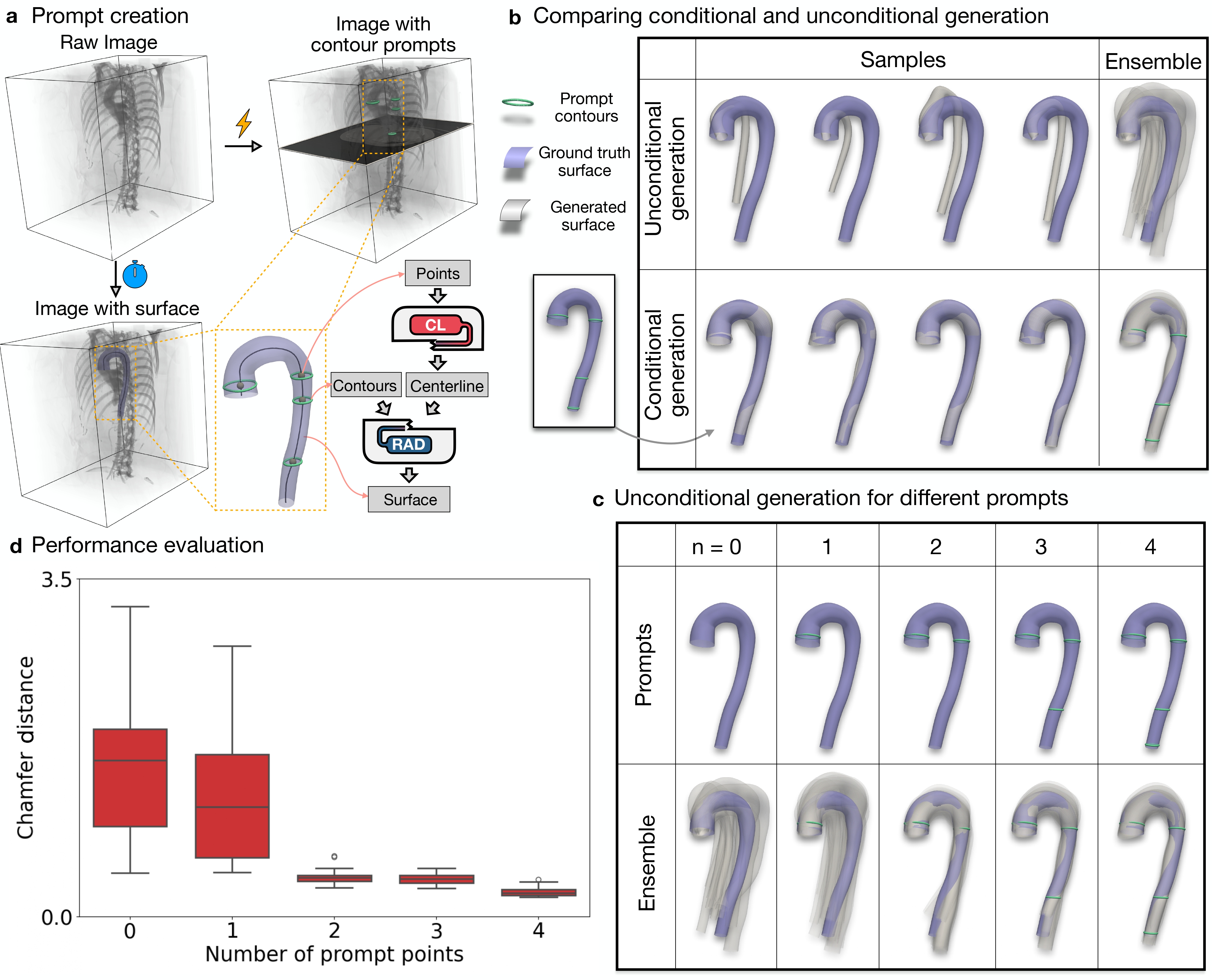}
    \caption{\textbf{a}, Schematic of HUG-VAS generating an aortic surface from sparse contour prompts. Instead of manually segmenting the full 3D surface from the raw image (left path), the user extracts contours from a few image slices (green loops). The centroid of each contour is used by the CL module to generate a centerline, which, together with the original contours, is passed to the RAD module to synthesize the full vessel surface. \textbf{b}, Comparison of unconditional (top) and conditional (bottom) generation. Conditional samples show reduced variability and stronger alignment with the ground truth. \textbf{c}, Ensembles generated with different numbers of contour prompts ($n=0$–4). Increasing the number of prompts progressively tightens the ensembles and lowers uncertainty. \textbf{d}, Quantitative evaluation using Chamfer distance. Both the mean and standard deviation decrease as more contour prompts are provided, reflecting improved fidelity and greater consistency. All results are shown on test cases not seen during training of the diffusion model, demonstrating its generalization capability.}
    \label{fig:cg2}
\end{figure}

We next evaluate conditional generation of vessel surfaces using sparse axial contour prompts segmented from image slices. In the example shown in Fig.~\ref{fig:cg2}, we select an unseen test case and visualize the extracted contours, ground-truth surface (purple), and centerline within the orange frame. Two contours are taken from the same slice, while the other two come from lower slices. Leveraging the hierarchical design of HUG-VAS, conditional generation proceeds through a two-stage DPS process: first, average point locations from the contour prompts are used to condition the centerline diffusion model; second, the generated centerline, together with the user-provided contour points, is passed into the radial-profile diffusion model to synthesize the surface. In practice, multiple centerlines are generated in parallel in the first stage, and multiple surfaces are reconstructed for each centerline in the second stage, introducing two layers of diversity into the final ensemble.

Figure~\ref{fig:cg2}b compares unconditional and conditional generations. While unconditional samples display substantial variability, conditional generation guided by four contour prompts produces surfaces that align tightly with the ground truth. A sensitivity analysis is shown in Fig.~\ref{fig:cg2}c: as the number of contour prompts increases, the uncertainty of the ensemble decreases, approaching negligible levels when four contours are provided. This trend is quantified in Fig.~\ref{fig:cg2}d, where the mean Chamfer distance drops from 1.57 (no prompts) to 0.26 (four prompts), and the standard deviation decreases from 0.73 to 0.052. These results confirm that user-provided contours act as strong constraints, concentrating the posterior distribution around the ground truth. 

In the context of image segmentation, this capability allows HUG-VAS to reconstruct surfaces directly from sparse contour inputs while also quantifying residual uncertainty, providing a practical route toward semi-automatic segmentation, reducing manual workload while maintaining high anatomical fidelity.

\subsection{Broader applications using Unconditional Generation}
While the conditional generation scenarios above highlight the potential of HUG-VAS for semi-automatic segmentation, its utility extends far beyond these examples, enabling clinically relevant applications that have not been addressed by traditional SSMs. 

\begin{figure}[t!]
    \centering
    \includegraphics[width=0.9\textwidth]{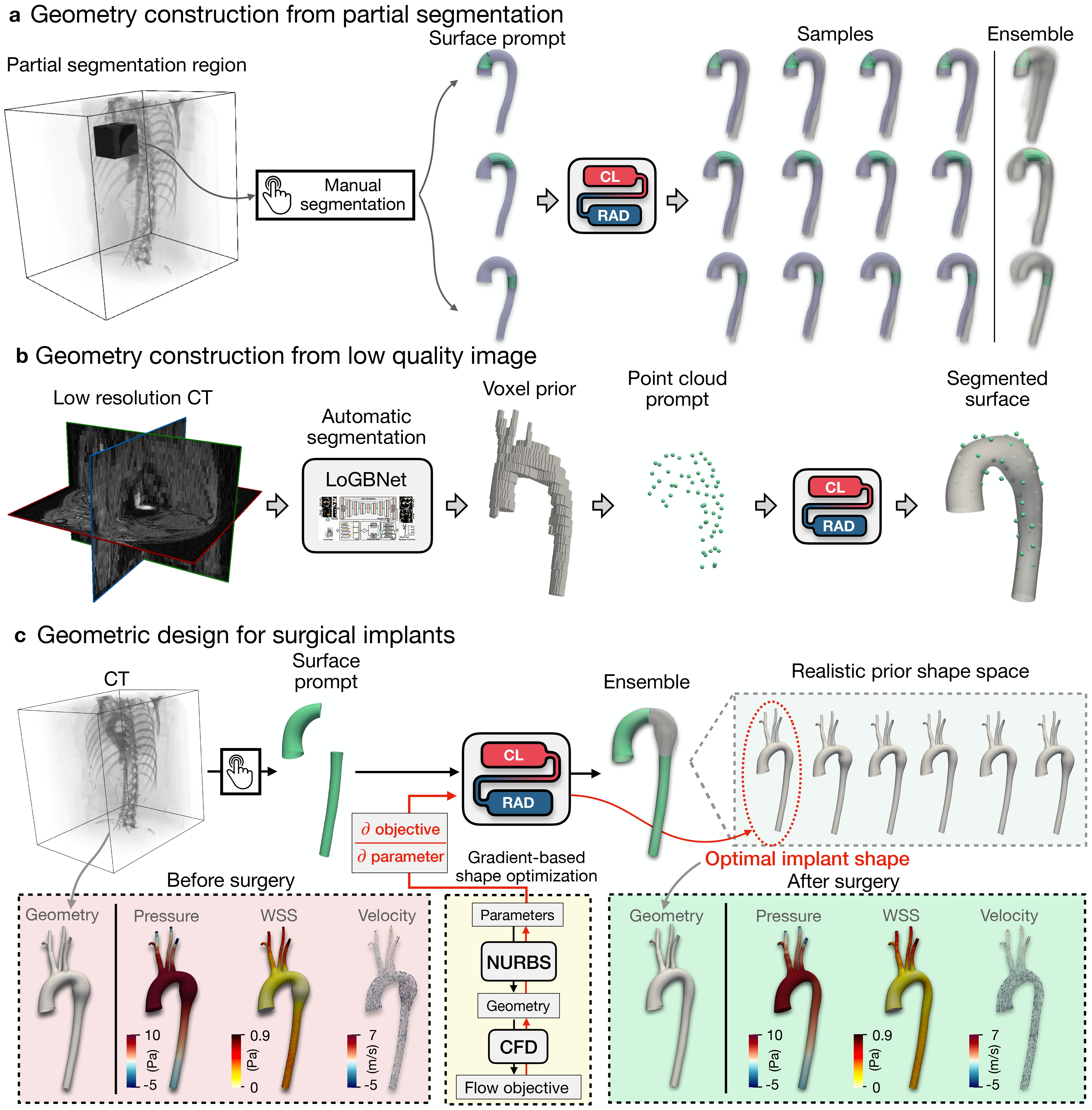}
    \caption{\textbf{a}, HUG-VAS reconstructs full aortic geometry from partial surface prompts manually segmented from medical images. \textbf{b}, For low-resolution CT scans, voxel-based predictions from LoGBNet are subsampled to create point prompts, from which HUG-VAS reconstructs smooth, high-quality surfaces. \textbf{c}, Application of HUG-VAS for surgical implant design in a thoracic aneurysm case. The healthy region (green) is segmented from the image and used as a conditional prompt to generate a library of personalized geometries spanning different aneurysm sizes, including fully healthy configurations. A differentiable pipeline (yellow box) is proposed to optimize this shape space with respect to hemodynamic objectives. Demonstrative pre- (left) and post- (right) intervention CFD results illustrate potential improvements in wall shear stress.}
    \label{fig:cg3}
\end{figure}

Figure~\ref{fig:cg3}a demonstrates the reconstruction of a complete aortic surface from partial segmentations extracted from limited image regions. This capability is especially valuable when scans suffer from poor signal quality or contain missing or corrupted regions. Users can isolate a high-quality patch, whether in the ascending aorta, arch, or descending aorta, and provide its surface segmentation (shown in green) as input. HUG-VAS then infers the full aortic geometry along with its posterior variability. Interestingly, ensembles generated from prompts in the ascending or descending regions exhibit greater variability than those from the arch, suggesting that the arch provides the most distinctive geometric cues in the dataset.

In addition to incomplete data, HUG-VAS is robust to image degradation such as low resolution. Figure~\ref{fig:cg3}b shows results on a rabbit CT scan with very poor resolution in the cranio-caudal direction. Conventional manual or automatic segmentation methods struggle to generate smooth, accurate results under these conditions. For example, using a well-trained auto-segmentation model like our previously developed LoGBNet~\cite{an2025hierarchical,du2025ai} can produce coarse, voxel-based predictions that are difficult to convert into clean surface meshes without manual post-processing, which often introduces geometric artifacts. In contrast, by subsampling the outermost voxel centers and using them as conditional inputs, HUG-VAS generates a high-quality, smooth surface reconstruction that adheres to the underlying voxel prediction. Remarkably, despite being trained exclusively on human aortas, HUG-VAS generalizes well to the rabbit case, highlighting strong cross-domain robustness.

Another promising application lies in the design of patient-specific surgical implants (Fig.~\ref{fig:cg3}c). We consider a thoracic aneurysm case where the clinical objective is to optimize the implant geometry. The workflow begins by segmenting only the healthy regions of the aorta (green), which are used as conditional prompts. HUG-VAS then generates an ensemble of plausible geometries spanning a prior shape space, including fully healthy configurations, represented compactly in the NURBS latent parameterization. This space can be integrated into a differentiable optimization pipeline (yellow box), where latent parameters map to geometry and then to CFD outputs. A clinical objective, such as reducing regions of abnormally low wall shear stress (WSS), can be defined, and gradients with respect to shape parameters are computed via AD. This enables gradient-based optimization to identify anatomically realistic geometries that meet hemodynamic targets. At the bottom of Fig.~\ref{fig:cg3}c, CFD simulations of both the aneurysmal anatomy and a representative optimized geometry illustrate the potential hemodynamic improvements, including reductions in low-WSS regions. While a full differentiable CFD pipeline is beyond the present scope, we outline this as a promising direction for future work (see Section~\ref{sec:discussion}).

\section{Discussion}
\label{sec:discussion}
\subsection{Exploring the generative space of HUG-VAS}
We have demonstrated both the conditional and unconditional generation capabilities of HUG-VAS. While the conditional scenarios introduced here are, to our knowledge, unprecedented in prior work, unconditional generation (i.e., synthesizing new anatomical geometries) has been explored before, though primarily in the context of single-channel vessels when applied to the aorta. Traditional approaches typically establish pointwise correspondences across surface meshes, apply PCA to the aligned dataset, and synthesize new samples by drawing PCA coefficients from a fitted multivariate Gaussian distribution.
To benchmark HUG-VAS against such methods, we implemented the baseline using PCA with Gaussian sampling across all five aortic branches. Surface correspondence was established by evaluating the NURBS representations at fixed streamwise and radial resolutions (e.g., $200\times80$ for the main aorta). For generation, we sampled the first 21 PCA coefficients from a multivariate  Gaussian distributions fitted from the training dataset.
In addition to this standard PCA baseline, we introduce a new ``decoupled PCA'' approach aligned with our hierarchical framework. Specifically, PCA was performed separately on latent encodings of centerlines and radial profiles, and independent multivariate Gaussian distributions were fitted for each. New samples were then generated by independently sampling centerline and radial coefficients. This strategy decouples centerline and radial variation during generation, enabling greater flexibility and diversity. 
We additionally include the widely used point voxel diffusion (PVD) model~\cite{zhou20213d} from the computer vision field as a baseline. However, it often fails to generate geometries with intact and topologically consistent surfaces (see details in Supplementary Note 4).

\begin{figure}[htp!]
    \centering
    \includegraphics[width=1.0\textwidth]{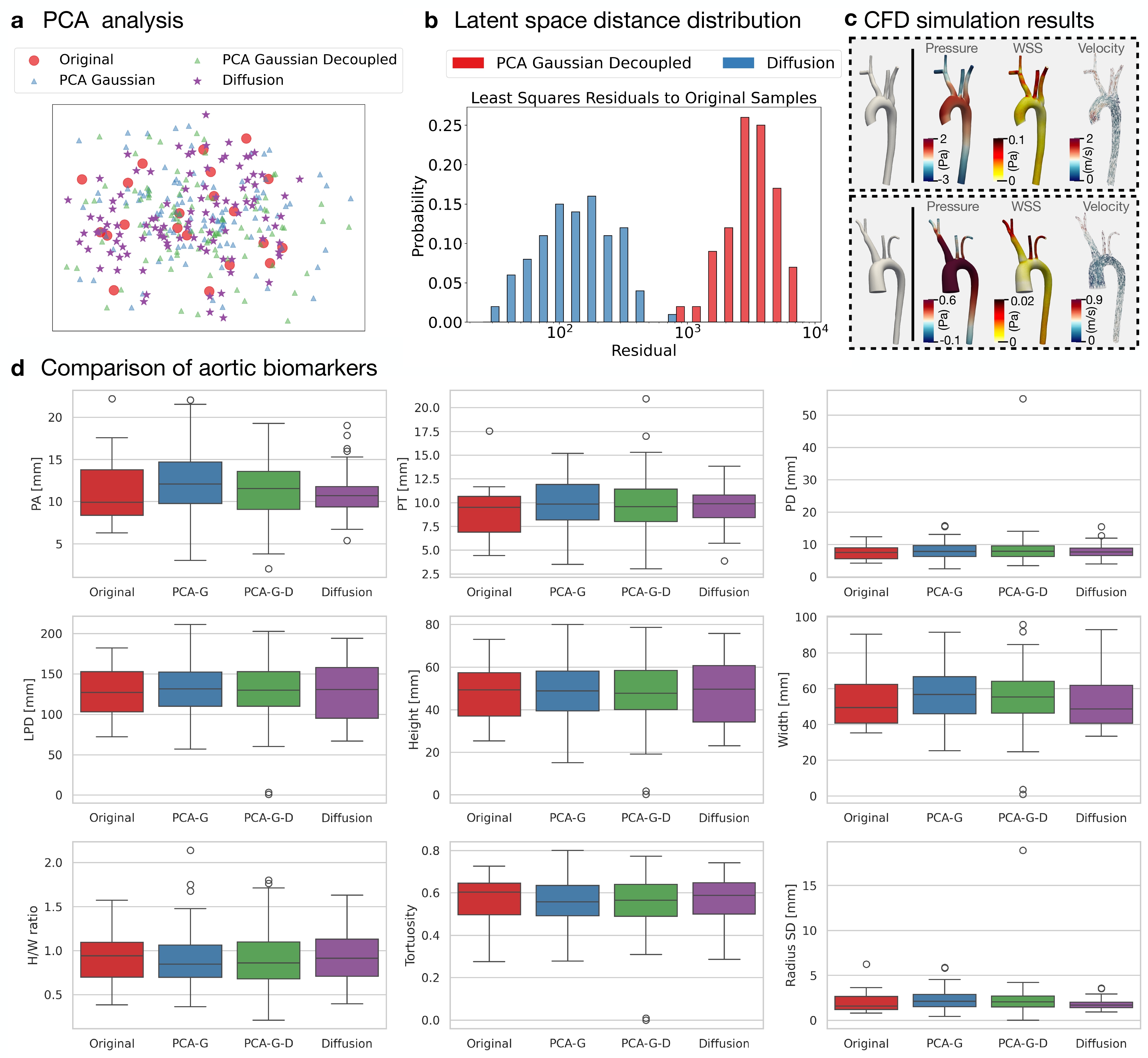}
    \caption{\textbf{a}, PCA projection of original and generated samples from three methods: PCA + Gaussian (PCA-G), PCA + Gaussian Decoupled (PCA-G-D), and HUG-VAS (Diffusion). \textbf{b}, Distribution of distances from generated samples to the original linear PCA latent space. PCA-G-D exhibits a unimodal distribution, while HUG-VAS shows a more complex, multimodal pattern. \textbf{c}, Representative CFD simulation results for generated geometries, visualizing pressure, wall shear stress (WSS), and velocity profiles. \textbf{d}, Comparison of aortic biomarkers across the original dataset and generated samples. While PCA-G and PCA-G-D yield broader variability, HUG-VAS (Diffusion) produces distributions that align more closely with the original cohort across key descriptors such as length (LPD), tortuosity, and radius variation.}
    \label{fig:pca}
\end{figure}

We generated 500 samples using HUG-VAS, standard PCA + Gaussian, and PCA + Gaussian (Decoupled). These samples, along with training data, were projected into the PCA latent space, and their distributions are shown in Fig.~\ref{fig:pca}a. The PCA-based methods exhibit broader coverage of the latent space than HUG-VAS, consistent with Romero et al.~\cite{romero2021clinically}, who reported greater variation from PCA with Gaussian sampling compared to PCA combined with a GAN. Although the PCA-based sampling methods can produce large shape variability, we found that it often yields visually implausible samples. Representative failed samples from the standard PCA + Gaussian, PCA + Gaussian (Decoupled), and PVD are presented in Supplementary Note 4. This non-physicality arises because PCA captures directions of maximal variance in point clouds without enforcing surface continuity or anatomical plausibility of the modes. Moreover, PCA constructs a linear subspace spanned by principal modes, which is fundamentally different from the nonlinear space learned by HUG-VAS. 
This difference is quantified in Fig.~\ref{fig:pca}b, which shows distances of generated samples to the PCA linear subspace derived from the PCA + Gaussian baseline. Both PCA + Gaussian (decoupled) and HUG-VAS yield nonzero distances, indicating that they explore generative spaces beyond the original linear PCA manifold. Interestingly, PCA + Gaussian (decoupled) produces a unimodal distance distribution, consistent with its underlying multivariate Gaussian assumption. In contrast, HUG-VAS yields a multimodal distribution, suggesting that it captures a more intricate generative space. This does not necessarily imply that HUG-VAS, or generative shape models more broadly, learns the ``true'' anatomical distribution, which remains unknown. Nevertheless, it is empirically reasonable to assume that the anatomical shape space is more complex than a unimodal distribution over a linear PCA subspace, and that HUG-VAS is better positioned to represent this complexity, particularly as larger training datasets become available.

\subsection{Simulation-ready mesh for seamless downstream CFD computations}
An important highlight of HUG-VAS is its ability to generate watertight, simulation-ready meshes that can be used directly in CFD solvers. While such compatibility is often taken for granted in statistical shape models, it represents a substantial improvement over auto-segmentation frameworks, where voxel-based outputs typically require extensive manual post-processing before they become CFD-compatible. To demonstrate the practicality of HUG-VAS, we performed CFD simulations on synthesized aortas, with two representative cases shown in Fig.~\ref{fig:pca}c. These simulations solve the steady incompressible Navier–Stokes equations under fixed-wall conditions (see Section~\ref{sec:meth}), confirming that HUG-VAS geometries can be deployed in CFD workflows without additional manual intervention.

\subsection{Evaluating sample realism with geometric biomarkers}
To assess the realism of HUG-VAS outputs, we computed nine classical geometric biomarkers (defined in Section~\ref{sec:meth}) for both the original dataset and synthetic datasets generated by the PCA baselines and HUG-VAS. The resulting distributions are shown in Fig.~\ref{fig:pca}d. Consistent with their broader latent coverage, ``PCA + Gaussian'' and ``PCA + Gaussian (decoupled)'' exhibit wider distributions across most biomarkers, reflecting higher shape variability. In contrast, HUG-VAS yields distributions that closely match those of the original dataset. Key biomarkers such as LPD (length from valve to proximal descending aorta), width, and tortuosity show nearly identical ranges and medians, highlighting strong fidelity in preserving global morphology and curvature. Overall, HUG-VAS generates synthetic samples that not only expand anatomical diversity but also faithfully reproduce the statistical characteristics of the training cohort, supporting their anatomical plausibility and downstream usability.

\subsection{Beyond shape synthesis: segmentation and surgical design with HUG-VAS}
\paragraph{Hierarchical architecture}  
The core design of HUG-VAS lies in its hierarchical architecture, which decouples the generation of centerlines and radial profiles. This two-stage strategy is inspired by traditional manual segmentation workflows in SimVascular, where users first define the centerline and then adjust cross-sectional radii under image guidance. In practice, centerline trajectories and radial profiles vary independently across patients, each following its own distribution while remaining anatomically correlated. Prior SSMs, however, typically enforce a strict one-to-one mapping between centerline and radius, thereby constraining variability in an overly rigid and unrealistic manner. By contrast, HUG-VAS generates radial profiles conditionally on the centerline, enabling diverse radius configurations for a fixed centerline shape. This conditional flexibility can be tuned via the scale factor $\gamma$ in the classifier-free guidance scheme: $\gamma=0$ allows the radial profile to vary independently, while increasing $\gamma$ progressively strengthens dependence on the centerline. At higher values (e.g., $\gamma=10$), the model tightly couples the two, effectively suppressing conditional variability and approaching a deterministic mapping (more details in Supplementary Note 5).

\paragraph{Semi-automatic segmentation}  
This decoupled variability naturally supports interactive semi-automatic segmentation—a hybrid paradigm that combines the accuracy of manual annotation with the efficiency of deep learning. The envisioned workflow proceeds as follows: (i) the user incrementally provides point prompts to generate centerline ensembles with progressively reduced uncertainty until convergence (Fig.~\ref{fig:cg1}d); (ii) contour prompts are supplied to generate vessel surfaces (Fig.~\ref{fig:cg2}d); and (iii) the resulting NURBS control points can be rendered and manually adjusted to correct residual misalignments with imaging data. To minimize effort in steps (i) and (ii), we propose an uncertainty-guided prompting strategy: new prompts are placed where ensemble variability is highest, thereby reducing uncertainty efficiently. The final NURBS-based editing step remains critical, as deep learning segmentation models inevitably inherit annotation artifacts from their training data. A graphical user interface (GUI) demonstration of the proposed semi-automatic segmentation workflow is presented in Supplementary Note 6 and Supplementary Video 1. A side-by-side comparison with the SimVascular workflow is provided in Supplementary Video 2, highlighting the efficiency and interactivity of HUG-VAS. 

\paragraph{Differentiable surgical design}  
HUG-VAS also supports full differentiability, making it directly integrable with differentiable CFD solvers and well-suited for many-query problems such as flow optimization and UQ. Figure~\ref{fig:cg3}c illustrates a pipeline for optimizing implantable devices in thoracic aneurysm cases. Two key distinctions separate this framework from conventional workflows.  
First, existing implant design often relies on trial-and-error iterations or parameter sweeps, where multiple configurations are manually tested to identify favorable outcomes~\cite{Hu2025FontanConduit,sahni2023quantitative}. This process is slow, requires substantial expertise, and lacks systematic convergence. In contrast, the differentiable framework of HUG-VAS enables gradient-based optimization, where shape parameters are efficiently updated using backpropagated gradients of a clinical objective, greatly accelerating convergence.  
Second, conditional ensembles generated from healthy vessel regions provide an anatomically informed prior shape space that spans physiologically realistic aneurysm variations. Optimization within this space ensures that final geometries remain consistent with known anatomy. Traditional workflows, by contrast, typically operate in heuristic parameter spaces that may converge to local optima, improving one flow metric but producing anatomically implausible geometries or introducing adverse hemodynamics elsewhere due to missing physiological constraints.

\subsection{Current limitations and paths forward}
Like most SSMs and auto-segmentation frameworks, HUG-VAS is inherently data-driven, which limits its ability to generalize to unseen pathologies. Our current dataset includes only thoracic aneurysm cases; as a result, the model cannot yet generate or segment aortas affected by other conditions such as coarctation or dissection. Expanding the training dataset to encompass a broader spectrum of pathologies is therefore an essential next step. In addition, coupling HUG-VAS with large language models (LLMs) could provide more intuitive, user-friendly control of the generation process through text-based prompts.

A second limitation lies in the current model configuration, which assumes a fixed number of branches and a predefined connectivity pattern. This design is well suited to the aorta but does not generalize to anatomies such as cerebral or coronary arteries, where branching topology varies considerably across individuals. Consequently, HUG-VAS is not directly applicable to vascular regions with more diverse or patient-specific topologies. Existing SSMs~\cite{Feldman2023VesselVAE,Feldman2025VesselGPT,Sinha2024TrIND,Kuipers2024ConditionalSetDiffusion,Prabhakar2024VesselGraphDD} focus explicitly on generating variable vascular tree structures. While HUG-VAS emphasizes high-fidelity surface detail, future integration with these topology-generation techniques could enable the synthesis of anatomically diverse vascular networks while retaining detailed surface fidelity. 


\section{Methodology}
\label{sec:meth}

\subsection{HUG-VAS framework}
\begin{figure}[htb!]
    \centering
    \includegraphics[width=1.0\textwidth]{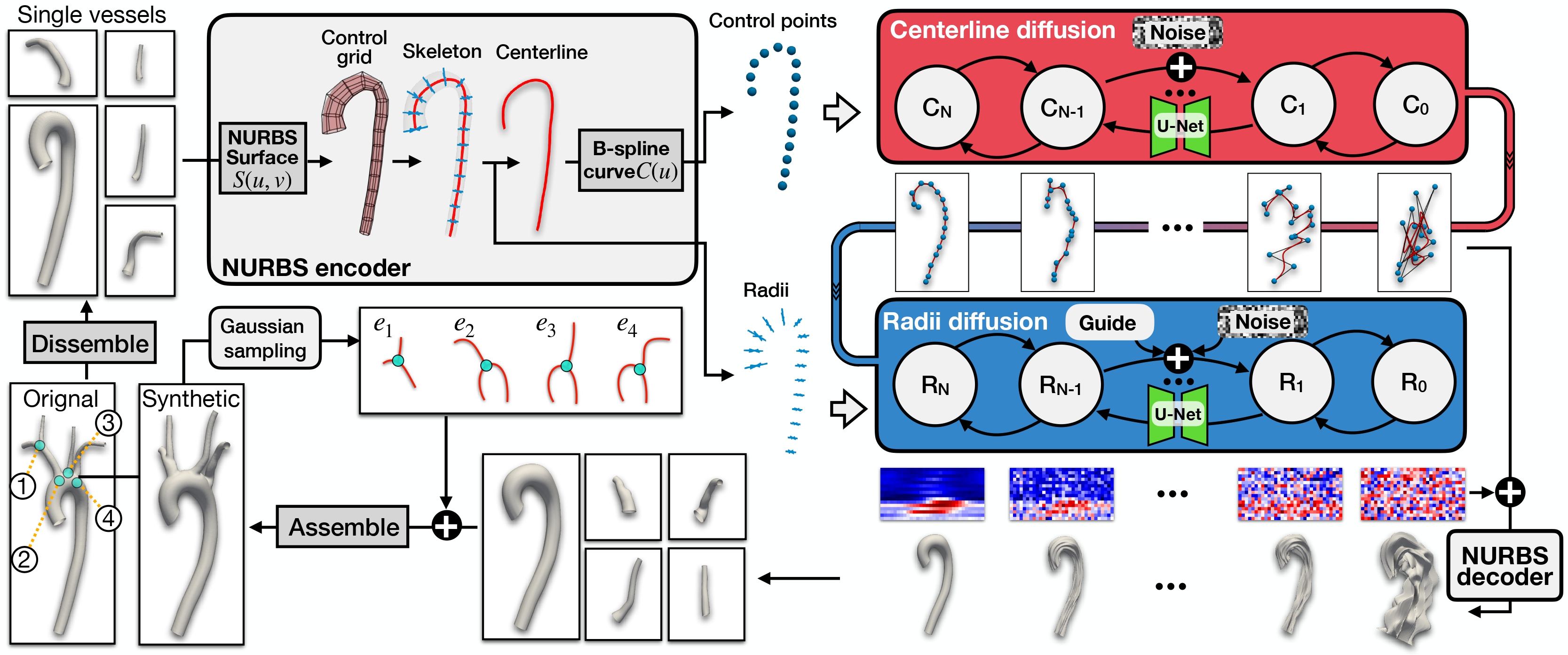}
    \caption{Schematic of the HUG-VAS hierarchical diffusion model with NURBS parameterization. Multi-branch aorta geometries are disassembled into individual vessels and encoded using NURBS. The centerline and radial profiles are modeled through two separate diffusion processes: centerline diffusion (red) and radius diffusion (blue), with the latter guided by the centerline. The generated control points and radii are decoded into full surfaces via the NURBS decoder and reassembled into multi-branch geometries. Gaussian sampling is used to generate new bifurcation locations during the assembly process.}
    \label{fig:method}
\end{figure}

HUG-VAS combines NURBS parameterization with hierarchical diffusion modeling to generate realistic aortic geometries in both unconditional and conditional modes. Figure~\ref{fig:method} illustrates the overall pipeline: (i) multi-branch aortas are decomposed into individual vessels; (ii) each vessel is encoded into latent representations of centerline control points and cross-sectional radii using a NURBS parameterization; (iii) separate diffusion models are trained on centerlines and radial profiles, with the latter conditioned on the former; (iv) unconditional or prompt-guided conditional generation is performed in the latent space; and (v) individual vessels are reconstructed via NURBS surfaces and reassembled into complete, watertight multi-branch geometries suitable for downstream CFD analysis.

\subsubsection{NURBS-based encoder and decoder}
We process the vessel geometries (see “NURBS encoder” in Fig.~\ref{fig:method}) into latent encodings through a series of procedural steps. First, vessel centerlines are extracted using the Vascular Modeling Toolkit (VMTK)~\cite{antiga2008image}. Specifically, each surface is extended at both ends with the \texttt{vmtkflowextensions} script (boundarynormal mode, extension ratio $1.1$) to address the common issue that the default \texttt{vmtkcenterlines} script produces shortened centerlines that do not reach vessel inlets or outlets. Centerlines are then computed with \texttt{vmtkcenterlines}.

Next, a NURBS fitting algorithm~\cite{piegl1997nurbs} is applied to extract the control points of the centerline. Given a centerline curve, we uniformly downsample it to $n$ points $\mathbf{Q}\in\mathbb{R}^{n \times 3}$, ordered from the distal to the proximal end. The B-spline curve $\mathcal{C}(u)$ is defined over a knot vector $\{u_0, u_1, \ldots, u_s\}$, where $s = n' + d_u + 1$, with $n' = n-1$ and $d_u$ the order of the basis functions. The mapping from control points $\mathbf{C} = \{\mathbf{c}_i\}_{i=0}^{n-1}$ to curve points $\mathbf{Q}$ is given by
\begin{equation}
    \mathbf{q}_k = \mathcal{C}(\bar{u}_k; \mathbf{C}) = \sum_{i=0}^{n-1} N_{i,d_u}(\bar{u}_k)\, \mathbf{c}_i,
\label{eq:nurbscurve}
\end{equation}
where $N_{i,d_u}$ is the $i$th basis function of order $d_u$, and $\bar{u}_k$ is the parameter corresponding to the $k$th curve point. The basis functions are defined recursively as
\begin{equation}
N_{i,d_u}(u) = 
\begin{cases}
1, & \text{if } u_i \leq u < u_{i+1}, \quad d_u = 0 \\
0, & \text{otherwise}, \quad d_u = 0 \\
\frac{u - u_i}{u_{i+d_u} - u_i} N_{i,d_u-1}(u) + \frac{u_{i+d_u+1} - u}{u_{i+d_u+1} - u_{i+1}} N_{i+1,d_u-1}(u), & \text{if } d_u > 0
\end{cases}
\label{eq:nurbs_basis}
\end{equation}
For curve fitting, we assign uniform parameters $\bar{u}_k$ over $[0,1]$ as,
\begin{equation}
\bar{u}_0 = 0, \quad \bar{u}_n = 1, \quad \bar{u}_k = \frac{k}{n}, \quad k = 1, \ldots, n-1,
\end{equation}
and determine the knot vector by
\begin{equation}
\begin{aligned}
& u_0 = \cdots = u_{d_u} = 0, \quad u_{s-d_u} = \cdots = u_s = 1, \\
& u_{j+d_u} = \frac{1}{d_u} \sum_{i = j}^{j + d_u - 1} \bar{u}_i, \quad j = 1, \ldots, n-d_u.
\end{aligned}
\end{equation}
These values are then used to evaluate the basis functions and solve the linear system in Eq.~\ref{eq:nurbscurve} for the control points $\mathbf{C}$.
With $\mathbf{C}$ obtained, tangent vectors $\mathbf{t}_k$ at each parameter location $\bar{u}_k$ are computed by,
\begin{equation}
\mathbf{t}_k = \frac{d\mathcal{C}(u)}{du} \Big|_{u = \bar{u}_k} = \sum_{i=0}^{n-1} \frac{dN_{i,d_u}(u)}{du} \Big|_{u = \bar{u}_k} \, \mathbf{c}_i.
\label{eq:tangent_vector}
\end{equation}
These tangents define the normal directions of cross-sectional planes on which radial profiles lie. To uniquely parameterize each profile, we assign an initial in-plane direction for the first cross-section as
\begin{equation}
\hat{\mathbf{w}}_0 = \frac{\mathbf{t}_0 \times \left(\mathbf{c}_n - \mathbf{c}_0\right)}{\|\mathbf{t}_0 \times (\mathbf{c}_n - \mathbf{c}_0)\|}.
\end{equation}
At each cross-section $i$, we generate a unit circular contour $\mathbf{K}^i = \{\mathbf{k}^i_j\}_{j=1}^o$ in the plane orthogonal to $\mathbf{t}_i$. Each contour has $o$ points ordered clockwise. Rotation-aligned reference vectors $\hat{\mathbf{w}}_i$ at subsequent sections are iteratively computed using Algorithm~\ref{alg:radial_alignment}.
\begin{algorithm}
\caption{Iterative Radial Vector Alignment Across Cross-Sections}
\label{alg:radial_alignment}
\begin{algorithmic}[1]
\State \textbf{Input:} Initial vector at the first cross section $\hat{\mathbf{w}}_0$; cross-section centers $\{\mathbf{q}_i\}_{i=0}^{n-1}$; contours $\{\mathbf{K}^i = \{\mathbf{k}^i_j\}_{j=1}^{o} \mid i = 0, \dots, n-1\}$
\State Compute radial vectors $\hat{\mathbf{r}}^0_j = \frac{\mathbf{k}^0_j - \mathbf{q}^0}{\|\mathbf{k}^0_j - \mathbf{q}^0\|}$ for all $j= 1, \dots, o$
\State Locate $j^{0,*} = \arg\max_j \left\{ \hat{\mathbf{r}}^0_j \cdot \mathbf{w}_0 | j =0, \dots, o-1 \right\}$
\For{$i = 1$ to $n-1$}
    \For{$l = 0$ to $o-1$}
        \State $d_l \gets \sum_{j=0}^{o} \left\| \mathbf{k}^{i-1}_j - \mathbf{k}^{i}_{(j + l) \bmod o} \right\|$
    \EndFor
    \State $l^* = \arg\min_l \{ d_0, d_1, \dots, d_{o-1} \}$
    \State $\hat{\mathbf{w}}_i = \frac{\mathbf{k}^i_{(j^{i-1,*} + l^*) \bmod o} - \mathbf{q}^i}{\left\| \mathbf{k}^i_{(j^{i-1,*} + l^*) \bmod o} - \mathbf{q}^i \right\|}$
    \State $j^{i,*} \gets (j^{i-1,*} + l^*) \bmod o$
\EndFor
\State \textbf{Output:} Aligned radial vectors $\{\hat{\mathbf{w}}_i\}_{i=0}^{n-1}$
\end{algorithmic}
\end{algorithm}
With the aligned reference axes $\{\hat{\mathbf{w}}_i\}_{i=0}^{n-1}$, we construct the full set of radial directions at each cross section by rotating $\hat{\mathbf{w}}_i$ counterclockwise in the plane orthogonal to the tangential vector $\mathbf{t}_i$. Specifically, the $j^{th}$ radial direction at the $i^{th}$ cross section is given by:
\begin{equation}
\hat{\mathbf{W}} = \left\{ \hat{\mathbf{w}}_{i,j} = \mathbf{R}(\mathbf{t}_i, j \delta\theta) \, \hat{\mathbf{w}}_i \;\middle|\; i = 0, \ldots, n-1;\; j = 0, \ldots, m - 1 \right\}, \quad \delta\theta = \frac{2\pi}{m}
\end{equation}
where $\mathbf{R}(\mathbf{t}_i, \delta\theta)$ denotes the 3D rotation matrix that rotates a vector around axis $\mathbf{t}_i$ by angle $\delta\theta$. We refer to the set $\{\mathbf{Q}, \hat{\mathbf{W}}\}$ as the skeleton of the vessel, as it serves as the structural basis for constructing the surface control points $\mathbf{S} = \{\mathbf{s}^i_j\}^{n-1,m-1}_{i =0, j = 0}$. Each surface control point is computed as:
\begin{equation}
\mathbf{s}^i_j = \xi\left(r^i_j; \mathbf{q}^i, \hat{\mathbf{w}}_{i,j} \right) = \mathbf{q}^i + r^i_j \cdot \hat{\mathbf{w}}_{i,j},  
\end{equation}
where $r^i_j$ is the $j^{\text{th}}$ radius at section $i$. The set $\mathbf{S}$ defines the surface polygon, which determines the vessel surface via the NURBS formulation:
\begin{equation}
\mathcal{S}(u,v; \mathbf{S}) = \frac{\sum_{i=0}^{n-1} \sum_{j=0}^{m-1} N_{i,d_u}(u) N_{j,d_v}(v) w_{i,j} \mathbf{\tilde{s}}^i_j}{\sum_{i=0}^{n-1} \sum_{j=0}^{m-1} N_{i,d_u}(u) N_{j,d_v}(v) w_{i,j}},
\label{eq:nurbs_surface}
\end{equation}
with $d_u=d_v=3$ and $w_{i,j}=1$. Control points $\mathbf{\tilde{s}}^i_j$ are padded to accommodate unclamped B-splines in the radial direction. The radial knot vector is:
\begin{equation}
\left\{-d_v*\delta u, -(d_v-1)*\delta u, ..., (m+1+d_v)*\delta u\right\}, \text{ where } \delta u = 1/m, 
\end{equation}
while the surface evaluation parameter $v$ remains within the standard range $[0,1]$.
The whole NURBS surface construction function $\mathcal{B}$ can be expanded as follows: 
\begin{equation}
\label{eq:nurbsforward}
\hat{\mathbf{v}} = \mathcal{B}\left(\mathbf{C}, \mathbf{R}\right) = \mathcal{S}(u,v; \mathbf{S}\left(\mathbf{C}, \mathbf{R}\right)) = \mathcal{S}(u,v; \xi\left(\mathbf{R}; \mathcal{C}\left( \bar{u};\mathbf{C}\right),\mathbf{W}\left(\mathbf{C}\right)\right))
\end{equation}

Given $\mathbf{C}$, the radial profile $\mathbf{R}$ is obtained by solving
\begin{equation}
\mathbf{R}^\ast = \arg\min_{\mathbf{R}} \mathcal{F}_{\text{chamfer}}\big(\mathcal{B}(\mathbf{C},\mathbf{R}),\,\mathbf{v}\big),
\label{eq:nurbsencoding}
\end{equation}
where $\mathcal{F}_{\text{chamfer}}$ is the bidirectional Chamfer distance between predicted and ground-truth surfaces. This completes the NURBS encoding, producing a consistent latent representation $\{\mathbf{C},\mathbf{R}^\ast\}$. Branch-specific discretization and mesh resolution are summarized in Table~\ref{tab:latentdimension}.
\begin{table}[h]
\centering
\caption{Latent representation dimensions and mesh resolution for each vascular branch.}
\label{tab:latentdimension}
\begin{tabular}{lccccc}
\toprule
 & Aorta & LCCA & LSA & RCCA & RSA \\
\midrule
Streamwise discretization        & 16  & 16  & 16  & 8   & 16  \\
Radial discretization            & 32  & 16  & 16  & 16  & 16  \\
Streamwise mesh resolution       & 200 & 120 & 120 & 60  & 120 \\
Radialwise mesh resolution       & 80  & 60  & 60  & 60  & 60  \\
\bottomrule
\end{tabular}
\end{table}

The NURBS decoder is simply the forward surface evaluation in Eq.~\ref{eq:nurbsforward}, mapping latent encodings $\{\mathbf{C},\mathbf{R}\}$ back to surface meshes $\mathbf{v}$. Notably, the resulting meshes are structured quadrilateral grids, aligned with the parameterized streamwise and radial directions.

\subsubsection{Unconditional generative modeling with hierarchical diffusion}

We learn the joint distribution of NURBS encodings in a \emph{hierarchical} fashion: a vanilla DDPM generates centerline control points $\mathbf{C}$, and a classifier-free guided diffusion generates radial profiles $\mathbf{R}$ conditioned on $\mathbf{C}$.

\paragraph{Centerline diffusion generation}
The forward (noising) process perturbs $\mathbf{C}$ with a fixed variance schedule $\{\beta_\tau\}_{\tau=1}^T$:
\begin{equation}
p(\mathbf{C}_\tau \mid \mathbf{C}_{\tau-1})=\mathcal{N}\!\left(\sqrt{1-\beta_\tau}\,\mathbf{C}_{\tau-1},\,\beta_\tau\mathbf{I}\right),\quad
\alpha_\tau=1-\beta_\tau,\;\bar{\alpha}_\tau=\textstyle\prod_{i=1}^\tau \alpha_i.
\end{equation}
This forward process defines a Markov chain from $\mathbf{C}_0$ to $\mathbf{C}_T$, where $\mathbf{C}_T$ becomes nearly isotropic Gaussian noise. The marginal distribution at step $\tau$ admits a closed form, leading to the reparameterization
\begin{equation}
\mathbf{C}_\tau=\sqrt{\bar{\alpha}_\tau}\,\mathbf{C}_0
+\sqrt{1-\bar{\alpha}_\tau}\,\boldsymbol{\epsilon},\qquad 
\boldsymbol{\epsilon}\sim\mathcal{N}(\mathbf{0},\mathbf{I}),
\label{eq:ddpm_marginal}
\end{equation}
enabling efficient training without simulating the entire Markov chain. The exact reverse process $p(\mathbf{C}_{\tau-1}\mid \mathbf{C}_\tau)$ is Gaussian but intractable to evaluate because it requires integration over the data distribution. Following Ho et al.~\cite{ho2020ddpm}, we parameterize the reverse mean using a neural network $\epsilon_\theta$ that predicts the forward noise in Eq.~\eqref{eq:ddpm_marginal}. The corresponding reverse mean estimator is
\begin{equation}
\boldsymbol{\mu}_\theta(\mathbf{C}_\tau,\tau)
=\frac{1}{\sqrt{\alpha_\tau}}\left(
\mathbf{C}_\tau-\frac{1-\alpha_\tau}{\sqrt{1-\bar{\alpha}_\tau}}\,
\boldsymbol{\epsilon}_\theta(\mathbf{C}_\tau,\tau)\right),
\end{equation}
with reverse variance fixed (or optionally learned) as $\sigma_\tau^2\mathbf{I}$. 

Training minimizes the variational lower bound (VLB), which reduces to a denoising score-matching objective. In practice we adopt the simplified form
\begin{equation}
\mathcal{L}_{\mathrm{simple}}
=\mathbb{E}_{\mathbf{C}_0,\tau,\boldsymbol{\epsilon}}
\Big\|
\boldsymbol{\epsilon}
-\boldsymbol{\epsilon}_\theta\!\left(\sqrt{\bar{\alpha}_\tau}\mathbf{C}_0
+\sqrt{1-\bar{\alpha}_\tau}\,\boldsymbol{\epsilon},\;\tau\right)
\Big\|_2^2,
\label{eq:ddpm_loss}
\end{equation}
which corresponds to an $\ell_2$ regression on the injected noise and is equivalent to score-matching under Gaussian assumptions,

At inference, centerline generation begins from $\mathbf{C}_T\sim\mathcal{N}(\mathbf{0},\mathbf{I})$ and proceeds by iteratively applying the learned reverse kernel,
\begin{equation}
\mathbf{C}_{\tau-1}=\boldsymbol{\mu}_\theta(\mathbf{C}_\tau,\tau)
+\sigma_\tau\,\boldsymbol{\xi},\qquad
\boldsymbol{\xi}\sim\mathcal{N}(\mathbf{0},\mathbf{I}),
\label{eq:ddpm_sampling}
\end{equation}
equivalent to integrating the reverse-time SDE in the score-based formulation~\cite{song2019score,song2021scorebased},
\begin{equation}
\mathbf{C}_{\tau-1}
= \frac{1}{\sqrt{\alpha_\tau}}
\Big(\mathbf{C}_\tau + (1-\alpha_\tau)\,\boldsymbol{\zeta}_\theta(\mathbf{C}_\tau,\tau)\Big)
+ \sigma_\tau\,\boldsymbol{\xi}, 
\end{equation}
where $\boldsymbol{\zeta}_\theta(\mathbf{C}_\tau,\tau) = -\frac{1}{\sqrt{1-\bar{\alpha}\tau}}\boldsymbol{\epsilon}_\theta(\mathbf{C}_\tau, \tau)  \approx\nabla_{\mathbf{C}_\tau}\log p(\mathbf{C}_\tau)$ is known as the score function.

\paragraph{Radius diffusion generation with classifier-free guidance}
The radial latent $\mathbf{R}$ is generated conditionally on the centerline $\mathbf{C}$ using a classifier-free guidance (CFG) scheme~\cite{ho2022classifierfree}. In this setting, a single network $\zeta_\theta$ is trained to predict scores for both unconditional and conditional distributions by randomly dropping the conditioning during training (with probability $p_{\text{drop}}$). This removes the need for a separate auxiliary classifier.
At inference, the conditional density is modified to
\begin{equation}
\tilde{p}_\theta(\mathbf{R}_0 \mid \mathbf{C})
\;\propto\; \frac{p_\theta(\mathbf{R}_0 \mid \mathbf{C})^{1+\gamma}}{p_\theta(\mathbf{R}_0)^\gamma},
\end{equation}
where $\gamma \geq 0$ controls the conditioning strength. In the score-based formulation, this corresponds to a guided score
\begin{equation}
\zeta_\theta^{\text{guided}}(\mathbf{R}_\tau,\tau\mid \mathbf{C})
=\zeta_\theta(\mathbf{R}_\tau,\tau)
+\gamma\Big(\zeta_\theta(\mathbf{R}_\tau,\tau\mid\mathbf{C})
-\zeta_\theta(\mathbf{R}_\tau,\tau)\Big),
\label{eq:r_guided_score}
\end{equation}
which interpolates between the unconditional score and the conditional score given $\mathbf{C}$.
The guided score \eqref{eq:r_guided_score} is then used in the reverse diffusion step, analogous to the centerline case:
\begin{equation}
\mathbf{R}_{\tau-1}
=\frac{1}{\sqrt{\alpha_\tau}}\Big(\mathbf{R}_\tau
+(1-\alpha_\tau)\,\zeta_\theta^{\text{guided}}(\mathbf{R}_\tau,\tau\mid\mathbf{C})\Big)
+\sigma_\tau\boldsymbol{\xi},
\qquad \boldsymbol{\xi}\sim\mathcal{N}(\mathbf{0},\mathbf{I}).
\label{eq:radii_sampling}
\end{equation}

This formulation preserves stochastic variability in $\mathbf{R}$ while explicitly allowing the conditioning signal from $\mathbf{C}$ to be scaled by $\gamma$. Small $\gamma$ encourages diverse radial profiles even for a fixed centerline, while large $\gamma$ enforces strong geometric coupling, yielding nearly deterministic cross-sectional shapes.

\subsubsection{Training-free conditional generation via DPS}

We perform zero-shot conditional generation with diffusion posterior sampling (DPS), applied sequentially to the centerline and radii models. Let $\mathbf{y}_C$ denote a centerline prompt (e.g., sparse 3D points) obtained from a differentiable forward operator $\mathcal{F}$, acting on the latent state. DPS samples from the posterior
\begin{equation}
p(\mathbf{C} \mid \mathbf{y}_C) \propto p(\mathbf{y}_C \mid \mathbf{C}) \, p(\mathbf{C}),
\end{equation}
where $p(\mathbf{C})$ is the generative prior learned by the diffusion model, and $p(\mathbf{y} \mid \mathbf{C})$ evaluates the likelihood of the conditions $\mathbf{y}$ given the current geometry via the forward mapping $\mathcal{F}(\mathbf{C})$. DPS modifies the sampling trajectory at each reverse step to align samples $\mathbf{C}$ with the gradient of the log posterior: 
\begin{equation}
\nabla_{\mathbf{C}_\tau} \log p(\mathbf{C}_\tau \mid \mathbf{y}_C) = \nabla_{\mathbf{C}_\tau} \log p(\mathbf{y}_C \mid \mathbf{C}_\tau) + \nabla_{\mathbf{C}_\tau} \log p(\mathbf{C}_\tau),
\end{equation}
where the second term is the score function of the unconditional generation process. The first term stems from the conditional information that can be expressed as:
\begin{equation}
p(\mathbf{y}_C \mid \mathbf{C}_\tau) = \int p(\mathbf{y}_C \mid \mathbf{C}_0, \mathbf{C}_\tau) \, p(\mathbf{C}_0 \mid \mathbf{C}_\tau) \, d\mathbf{C}_0 = \mathbb{E}_{\mathbf{C}_0 \sim p(\mathbf{C}_0 \mid \mathbf{C}_\tau)} \left[ p(\mathbf{y}_C \mid \mathbf{C}_0) \right],
\end{equation}
which can be approximated by:
\begin{equation}
p(\mathbf{y}_C \mid \mathbf{C}_\tau) = \mathbb{E}_{\mathbf{C}_0 \sim p(\mathbf{C}0 \mid \mathbf{C}\tau)}[p \mid \mathbf{C}_0]\approx p\left(\mathbf{y}_C \mid \mathbb{E}_{\mathbf{C}_0 \sim p(\mathbf{C}_0 \mid \mathbf{C}_\tau)}[\mathbf{C}_0\mid\mathbf{C}_\tau] \right).
\end{equation}
The approximation error is theoretically bounded by the Jensen gap. Therefore, the gradient of the conditional log-likelihood can be approximated using the expected posterior mean:
\begin{equation}
\nabla_{\mathbf{C}_\tau} \log p(\mathbf{y}_{C} \mid \mathbf{C}_\tau)
\approx
\nabla{\mathbf{C}_\tau} \log p\left(\mathbf{y}_{C} \mid \mathbb{E}[\mathbf{C}_0 \mid \mathbf{C}_\tau]\right)
\end{equation}
where the posterior mean $\hat{\mathbf{C}}_0 = \mathbb{E}[\mathbf{C}_0 \mid \mathbf{C}_\tau]$ can be computed as:
\begin{equation}
\hat{\mathbf{C}}_0 = 
\frac{1}{\sqrt{\bar{\alpha}_\tau}}
\left(
\mathbf{C}\tau + (1 - \bar{\alpha}_\tau) \nabla_{\mathbf{C}_\tau} \log p(\mathbf{C}_\tau)
\right), 
\end{equation}
where the Stein score $\nabla_{\mathbf{C}_\tau} \log p(\mathbf{C}_\tau)$ is already learned during the unconditional training, leading to approximation of $\hat{\mathbf{C}}_0$ as: 
\begin{equation}
\hat{\mathbf{C}}_0 \approx \hat{\mathbf{C}}_0^*(\mathbf{C}_\tau, \tau; \boldsymbol{\theta}^*) = \frac{1}{\sqrt{\bar{\alpha}_\tau}} \left( \mathbf{C}_\tau + (1 - \bar{\alpha}_\tau) \mathbf{\zeta}_{\boldsymbol{\theta}^*}(\mathbf{C}_\tau, \tau; \boldsymbol{\theta}^*) \right).
\end{equation}
The nonlinear mapping $\mathcal{F}$ is composed of the B-spline function $\mathcal{C}$ and the observation function $\mathcal{O}$, resulting in the approximated likelihood distribution:
\begin{equation}
p(\mathbf{y}_C \mid \mathbf{C}_0) \approx p(\mathbf{y}_C \mid \mathbf{C}_0^*) \sim \mathcal{N} \left( \mathcal{O}(\mathcal{C}(\hat{\mathbf{C}}_0^*(\mathbf{C}_\tau, \tau; \boldsymbol{\theta}^*))), \sigma^2\mathbf{I} \right).
\end{equation}
Differentiating the log-likelihood with respect to $\mathbf{C}_\tau$ yields the following approximation:
\begin{equation}
\begin{aligned}
\nabla_{\mathbf{C}_\tau} \log p(\mathbf{y}_C \mid \mathbf{C}_\tau)
&\approx \nabla_{\mathbf{C}_\tau} \log p_{\boldsymbol{\theta}^*}(\mathbf{y}_C \mid \mathbf{C}_\tau) \\
&= -\frac{2}{\sigma_c^2} \left( \mathbf{y}_C - \mathcal{O}(\mathcal{C}({\mathbf{C}}_0^*) \right)
\frac{\partial \mathcal{O}(\mathcal{C}({\mathbf{C}}_0^*))}{\partial \mathcal{C}(\hat{\mathbf{C}}_0^*)} 
\frac{\partial \mathcal{C}({\mathbf{C}}_0^*)}{\partial \mathbf{C}_0^*}
\frac{\partial \mathbf{C}_0^*(\mathbf{C}_\tau, \tau; \boldsymbol{\theta}^*)}{\partial \mathbf{C}_\tau}, 
\end{aligned}
\end{equation}
which can be computed via AD. Finally, the guided diffusion score under DPS take the form: 
\begin{equation}
\begin{aligned}
\nabla_{\mathbf{C}_\tau} \log p(\mathbf{C}_\tau \mid \mathbf{y}_C)
&\approx 
\nabla_{\mathbf{C}_\tau} \log p_{\boldsymbol{\theta}^*}(\mathbf{y}_C \mid \mathbf{C}_\tau) +
\nabla_{\mathbf{C}_\tau} \log p_{\boldsymbol{\theta}^*}(\mathbf{C}_\tau)\\
&= 
\underbrace{\nabla_{\mathbf{C}_\tau} \log p_{\boldsymbol{\theta}^*}(\mathbf{y}_C \mid \mathbf{C}_\tau)}_{\text{DPS guidance}} +
\underbrace{\mathbf{\zeta}_{\boldsymbol{\theta}^*}(\mathbf{C}_\tau, \tau)}_{\text{Unconditional score}} \\
&= 
\mathbf{\zeta}_{\boldsymbol{\theta}^*}^{\text{guided}}(\mathbf{y}_C, \mathbf{C}_\tau, \tau; \boldsymbol{\theta}^*).
\label{eq:centerline_cg}
\end{aligned}
\end{equation}

For the radii diffusion, the DPS-guided score function adopts a similar form, with an additional modification incorporating guidance from the centerline: 
\begin{equation}
\begin{aligned}
\nabla_{\mathbf{R}_\tau} \log p(\mathbf{R}_\tau \mid \mathbf{y}_R, \mathbf{C}) 
&=\boldsymbol{\zeta}^{\text{guided}}{\boldsymbol{\theta}^*}(\mathbf{y}_R, \mathbf{R}_\tau, \tau; \mathbf{C})
\\
&= \underbrace{\boldsymbol{\zeta}_{\boldsymbol{\theta}^*}(\mathbf{R}_\tau, \tau)}_{\text{unconditional score}} 
+ \underbrace{\nabla_{\mathbf{R}_\tau} \log p_{\boldsymbol{\theta}^*}(\mathbf{y}_R \mid \mathbf{R}_\tau)}_{\text{DPS observation guidance}} \\
&\quad + \underbrace{\gamma \left( \boldsymbol{\zeta}_{\boldsymbol{\theta}^*}(\mathbf{R}_\tau, \tau \mid \mathbf{C}) - \boldsymbol{\zeta}_{\boldsymbol{\theta}^*}(\mathbf{R}_\tau, \tau) \right)}_{\text{centerline CFG}}. \label{eq:radii_cg}
\end{aligned}
\end{equation}
Details of the denoising neural network architecture and the procedures for both unconditional and DPS-guided conditional generation are provided in the Supplementary Note 1.

\subsubsection{Statistical assembly of multi-branch vascular geometries}
\begin{figure}[htb!]
    \centering
    \includegraphics[width=0.9\textwidth]{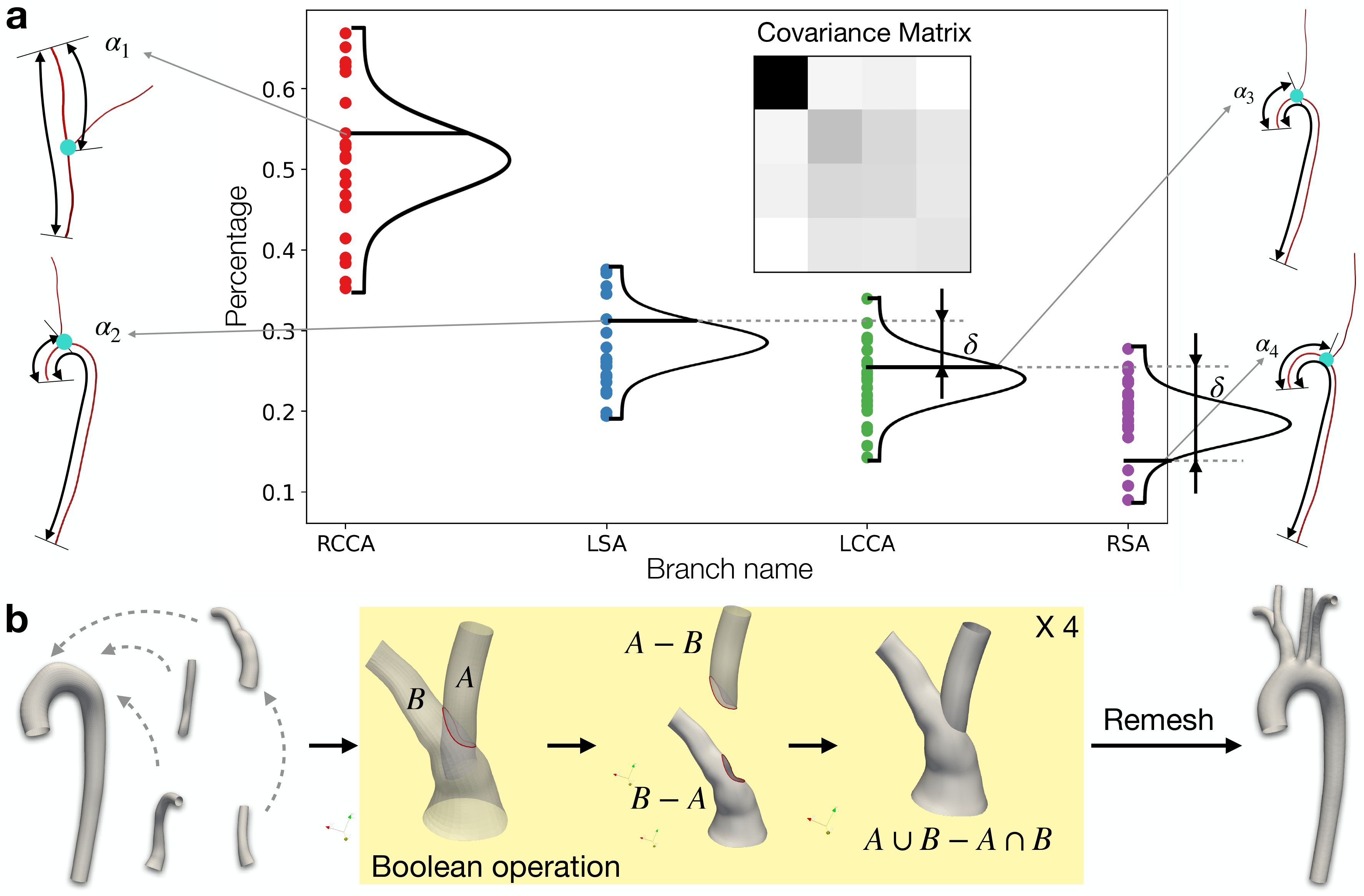}
    \caption{{\textbf{a}, Statistical analysis of branching locations across four supra-aortic vessels: RCCA, LSA, LCCA, and RSA. Each point represents the normalized location of a bifurcation relative to the arc-length of the parent vessel. The covariance matrix (top) captures interdependence among branch positions across the dataset. \textbf{b}, Boolean operations are then performed to merge the bifurcations, followed by remeshing to produce the final simulation-ready geometry.}}
    \label{fig:boolean}
\end{figure}

The multi-branch aorta configuration considered in this work follows a fixed topology with four bifurcations ($f=4$) and five vessels ($b=5$: main aorta, LCCA, LSA, RCCA, and RSA). Instead of learning topology, we parameterize only the branch locations $\mathbf{E}=\{e_i\}_{i=1}^f$, where each $e_i$ specifies the normalized arc-length position of a bifurcation along its parent centerline. From the training dataset $\{\mathbf{E}^j\}_{j=1}^N$, we fit a multivariate Gaussian distribution that captures both the marginal statistics of individual branch positions and their cross-dependencies (Fig.~\ref{fig:boolean}a). During synthesis, we first sample $\hat{\mathbf{E}}$ from this distribution and generate the five vessel segments independently via the hierarchical diffusion model. The individual vessels are then merged into a coherent multi-branch structure by performing Boolean union operations at the sampled bifurcation locations, followed by local Laplacian smoothing to ensure watertight junctions and surface continuity (Fig.~\ref{fig:boolean}b). This procedure yields anatomically plausible assemblies that preserve the learned branch-location statistics while producing CFD-ready meshes.

\subsection{Dataset, training, and testing settings}

\subsubsection{Dataset Preparation}

The primary dataset was obtained from the Vascular Model Repository (VMR)~\cite{Wilson2013}, from which we selected 21 human thoracic aorta cases covering a range of anatomies, including healthy subjects, post-Fontan congenital heart disease, and thoracic aneurysms. All data were acquired from MRA scans, and the aorta and supra-aortic branches were manually segmented using SimVascular by trained annotators following standard workflows. To evaluate cross-modality generalization, we included one CT case from the Multicenter Aortic Vessel Tracking (AVT) dataset~\cite{Radl2022}, which was manually segmented and used in Figs.~\ref{fig:cg1}, \ref{fig:cg2}, and \ref{fig:cg3}a. Additional external tests included (i) one unseen rabbit CT case from the VMR (Fig.~\ref{fig:cg2}b) to assess cross-species generalization, and (ii) withheld human cases from VMR (Fig.~\ref{fig:cg2}c). Prior to training, all segmented vessels were grouped by branch type (aorta, LCCA, LSA, RCCA, RSA). Each branch was centered by subtracting the mean of its point coordinates to remove translational offsets, while preserving relative scale and orientation.

\subsubsection{Training and testing settings}
We trained five hierarchical diffusion models, one for each vascular branch. Each model comprises two components: (i) a DDPM for centerline generation and (ii) a classifier-free guided diffusion for radial profile generation, both trained independently on NVIDIA RTX 4090 GPUs. Training required approximately 20 minutes per component. We used the Adam optimizer with a learning rate of $8 \times 10^{-5}$ and a batch size of 110. All latent encodings were normalized to $[0,1]$ prior to training.  

During conditional generation, DPS was used to incorporate prompt information. The guidance term is defined by the differentiable Chamfer loss between generated samples and prompts (e.g., centerline points, surface contours), computed bidirectionally between two point clouds $\mathbf{X} = \{\mathbf{x}_i\}_{i=1}^{N_x}$ and $\mathbf{G} = \{\mathbf{g}_j\}_{j=1}^{N_g}$:
\begin{equation}
\mathcal{F}_{\text{chamfer}}(\mathbf{X}, \mathbf{G}) = \frac{1}{N_x} \sum_{\mathbf{x} \in \mathbf{X}} \min_{\mathbf{g} \in \mathbf{G}} \|\mathbf{x} - \mathbf{g}\|_2^2 + \frac{1}{N_g} \sum_{\mathbf{g} \in \mathbf{G}} \min_{\mathbf{x} \in \mathbf{X}} \|\mathbf{g} - \mathbf{x}\|_2^2,
\end{equation}
which we implemented in a fully differentiable manner.

For generation, unconditional and conditional sampling both required $\sim$2 minutes per case. The denoising process supports batch inference, and we used a batch size of 50 for both centerline and radii synthesis. All ground-truth geometries used in the test set were manually segmented following standard procedures in SimVascular.

\subsection{PCA-based SSM baselines}
\paragraph{PCA + Gaussian} 
As a baseline SSM, we adopt the PCA + Gaussian sampling approach proposed by Romero et al.~\cite{romero2021clinically}. We first evaluate the NURBS surface for all training samples to generate point-to-point corresponded surface meshes. PCA is then performed to obtain orthogonal modes ranked by their explained variance. We subsequently fit a multivariate Gaussian distribution over the PCA coefficients: Given a dataset of coefficients $\{\mathbf{a}^i\}_{i=1}^{N}$, where $\mathbf{a}^i = (a^i_1, \dots, a^i_{N_d})$ is the feature vector of dimension $N_d$, the Gaussian distribution $\mathcal{N}(\boldsymbol{\mu}, \boldsymbol{\Sigma})$ is parameterized by:
\begin{equation}
\left\{
\begin{aligned}
\boldsymbol{\mu^*} &= \bar{\mathbf{a}} =  \frac{1}{K_0} \sum_{i=1}^{K_0} \mathbf{a}^i, \\
\boldsymbol{\Sigma^*} &= \frac{1}{K_0 - 1} \sum_{i=1}^{K_0} (\mathbf{a}^i - \bar{\mathbf{a}})(\mathbf{a}^i - \bar{\mathbf{a}})^\top,
\label{eq:multiGaussian}
\end{aligned}
\right.
\end{equation}
where $\boldsymbol{\mu^*}$ and $\boldsymbol{\Sigma}^*$ are the fitted mean and covariance matrix of the multivariate Gaussian distribution.
With this fitted distribution, new samples can be synthesized by drawing a random set of PCA coefficients and reconstructing the shape using the principal modes:
\begin{equation}
\mathbf{v}^{\text{new}} = \bar{\mathbf{v}} + \mathbf{U} \mathbf{a}^{\text{new}},
\end{equation}
where $\bar{\mathbf{v}}$ is the mean surface, $\mathbf{U} \in \mathbb{R}^{N_v \times N_d}$ is the PCA basis matrix (with each column representing a principal mode), and $\mathbf{a}^{\text{new}} \sim \mathcal{N}(\boldsymbol{\mu}^*, \boldsymbol{\Sigma}^*)$ is a new set of latent coefficients sampled from the learned Gaussian distribution.

\paragraph{PCA + Gaussian decoupled}
To align the baseline with the hierarchical architecture of HUG-VAS, we introduce a variant in which the centerline and radial profile are modeled independently. Specifically, we perform PCA separately on centerline control points and radial profiles, fitting two independent Gaussians:
\[
p_{\mathbf{C}}=\mathcal{N}(\boldsymbol{\mu}_C,\boldsymbol{\Sigma}_C),\qquad
p_{\mathbf{R}}=\mathcal{N}(\boldsymbol{\mu}_R,\boldsymbol{\Sigma}_R).
\]
At generation time, we sample $\mathbf{a}_C^{\text{new}}\sim p_{\mathbf{C}}$ and $\mathbf{a}_R^{\text{new}}\sim p_{\mathbf{R}}$, and reconstruct the geometry by decoding with the NURBS operator:
\begin{equation}
\mathbf{v}^{\text{new}}=\mathcal{B}\!\left(
\bar{\mathbf{C}}+\mathbf{U}_C\mathbf{a}_C^{\text{new}},\;
\bar{\mathbf{R}}+\mathbf{U}_R\mathbf{a}_R^{\text{new}}
\right),
\end{equation}
where $\mathbf{U}_C,\mathbf{U}_R$ are the PCA bases for centerline and radial profiles, respectively. This “Gaussian decoupling” strategy allows centerline and radius variation to be sampled independently, providing a closer analogue to the hierarchical generative process in HUG-VAS.

\subsection{Evaluation}

\subsubsection{Biomarkers}
To quantitatively characterize aortic morphological variability, we extract a set of geometric biomarkers from each centerline. These biomarkers capture both local vessel dimensions and global shape features. Specifically, we measure the radii at three key anatomical landmarks: the ascending aorta near the sinotubular junction (PA), the top of the aortic arch (PT), and the descending thoracic aorta at the location opposite to PA (PD). The centerline length from the aortic valve to PD (LPD) serves as a surrogate for vessel elongation. Arch height ($h$) and width ($w$) define the overall arch curvature, from which we compute derived indices such as the height-to-width ratio ($h/w$) and the tortuosity index (tor), defined as
\[
tor = 1-\frac{w}{LPD} .
\]
In addition, we compute the standard deviation of cross-sectional radii along the centerline (Radius SD), which reflects localized dilatation or uneven vessel remodeling. Collectively, these biomarkers provide interpretable, clinically relevant features for assessing generated geometries against real anatomical data.

\subsubsection{Downstream CFD settings}
To evaluate the CFD compatibility of the reconstructed aorta meshes, we conduct steady-state simulations using OpenFOAM~\cite{OpenFOAM}, a widely used open-source CFD solver. The incompressible Navier-Stokes equations are solved using the SIMPLE algorithm:
\begin{equation}
\left\{
\begin{aligned}
\nabla \cdot \mathbf{u} &= 0, \\
(\mathbf{u} \cdot \nabla)\mathbf{u} &= -\nabla p + \nu \nabla^2 \mathbf{u},
\end{aligned}
\right.
\label{eq:ns}
\end{equation}
where $\mathbf{u}$ denotes the velocity field, $p$ is the pressure, and $\nu$ is the kinematic viscosity. Physiological boundary conditions are applied, including a parabolic velocity profile at the inlet (with a peak velocity of 1 m/s), fixed pressure or zero-gradient conditions at the outlets, and no-slip conditions on vessel walls. The solver uses second-order Gauss linear schemes for spatial discretization of gradients, divergence, and Laplacian terms within a steady-state formulation. 
Each volumetric mesh contains approximately one million tetrahedral cells, ensuring grid resolution and independence. Simulations are run on a single CPU core, with typical convergence in $\sim$30 minutes per case. Key hemodynamic outputs include velocity fields, pressure distributions, and wall shear stress (WSS), which are extracted for functional assessment of both real and generated anatomies.

 \section*{Data availability}
The datasets generated and/or analysed during the current study are available in the Zenodo repository, https://doi.org/10.5281/zenodo.18302984.

 \section*{Acknowledgment}
 This work was supported by the National Science Foundation under grant OAC-2047127 and the National Institutes of Health under award number 1R01HL177814 (J.X.~W.).

 \section*{Author Contributions}
 PD conceived the study, conducted experiments, performed analysis, and drafted the manuscript, including figures and supplementary materials.
MX conducted experiments, contributed to analysis, and assisted with manuscript preparation. 
XZ conducted experiments and contributed to figure preparation.
JW conceived and supervised the study, provided critical intellectual input, and revised the manuscript. 
All authors read and approved the final manuscript.

 \section*{Ethics declarations}
\subsection*{Competing interests}

The authors declare no competing interests.





\bibliographystyle{elsarticle-num}

\begin{thebibliography}{100}
	\expandafter\ifx\csname url\endcsname\relax
	\def\url#1{\texttt{#1}}\fi
	\expandafter\ifx\csname urlprefix\endcsname\relax\def\urlprefix{URL }\fi
	\expandafter\ifx\csname href\endcsname\relax
	\def\href#1#2{#2} \def\path#1{#1}\fi
	
	\bibitem{nabel2003cardiovascular}
	E.~G. Nabel, Cardiovascular disease, New England Journal of Medicine 349~(1) (2003) 60--72.
	
	\bibitem{sel2024building}
	K.~Sel, D.~Osman, F.~Zare, S.~Masoumi~Shahrbabak, L.~Brattain, J.-O. Hahn, O.~T. Inan, R.~Mukkamala, J.~Palmer, D.~Paydarfar, et~al., Building digital twins for cardiovascular health: from principles to clinical impact, Journal of the American Heart Association 13~(19) (2024) e031981.
	
	\bibitem{taylor2009patient}
	C.~A. Taylor, C.~Figueroa, Patient-specific modeling of cardiovascular mechanics, Annual review of biomedical engineering 11~(1) (2009) 109--134.
	
	\bibitem{coorey2022health}
	G.~Coorey, G.~A. Figtree, D.~F. Fletcher, V.~J. Snelson, S.~T. Vernon, D.~Winlaw, S.~M. Grieve, A.~McEwan, J.~Y.~H. Yang, P.~Qian, et~al., The health digital twin to tackle cardiovascular disease—a review of an emerging interdisciplinary field, NPJ digital medicine 5~(1) (2022) 126.
	
	\bibitem{marcos2021applications}
	L.~J. Marcos-Zambrano, K.~Karaduzovic-Hadziabdic, T.~Loncar~Turukalo, P.~Przymus, V.~Trajkovik, O.~Aasmets, M.~Berland, A.~Gruca, J.~Hasic, K.~Hron, et~al., Applications of machine learning in human microbiome studies: a review on feature selection, biomarker identification, disease prediction and treatment, Frontiers in microbiology 12 (2021) 634511.
	
	\bibitem{bridio2023generation}
	S.~Bridio, G.~Luraghi, A.~Ramella, J.~F. Rodriguez~Matas, G.~Dubini, C.~A. Luisi, M.~Neidlin, P.~Konduri, N.~Arrarte~Terreros, H.~A. Marquering, et~al., Generation of a virtual cohort of patients for in silico trials of acute ischemic stroke treatments, Applied Sciences 13~(18) (2023) 10074.
	
	\bibitem{vukicevic2017cardiovascular}
	M.~Vukicevic, B.~Mosadegh, J.~K. Min, S.~H. Little, Cardiovascular 3d printing: From clinical applications to bench research, Journal of the American College of Cardiology 70~(25) (2017) 2975--2984.
	
	\bibitem{durrleman2009statistical}
	S.~Durrleman, X.~Pennec, A.~Trouv{\'e}, N.~Ayache, Statistical models of sets of curves and surfaces based on currents, Medical image analysis 13~(5) (2009) 793--808.
	
	\bibitem{twining2005unified}
	C.~J. Twining, T.~Cootes, S.~Marsland, V.~Petrovic, R.~Schestowitz, C.~J. Taylor, A unified information-theoretic approach to groupwise non-rigid registration and model building, in: Information Processing in Medical Imaging: 19th International Conference, IPMI 2005, Glenwood Springs, CO, USA, July 10-15, 2005. Proceedings 19, Springer, 2005, pp. 1--14.
	
	\bibitem{leventon2000level}
	M.~E. Leventon, O.~Faugeras, W.~E.~L. Grimson, W.~M. Wells, Level set based segmentation with intensity and curvature priors, in: Proceedings IEEE Workshop on Mathematical Methods in Biomedical Image Analysis. MMBIA-2000 (Cat. No. PR00737), IEEE, 2000, pp. 4--11.
	
	\bibitem{masala2013improved}
	G.~L. Masala, B.~Golosio, P.~Oliva, An improved marching cube algorithm for 3d data segmentation, Computer Physics Communications 184~(3) (2013) 777--782.
	
	\bibitem{zhang2007patient}
	Y.~Zhang, Y.~Bazilevs, S.~Goswami, C.~L. Bajaj, T.~J. Hughes, Patient-specific vascular nurbs modeling for isogeometric analysis of blood flow, Computer methods in applied mechanics and engineering 196~(29-30) (2007) 2943--2959.
	
	\bibitem{updegrove2017simvascular}
	A.~Updegrove, N.~M. Wilson, J.~Merkow, H.~Lan, A.~L. Marsden, S.~C. Shadden, Simvascular: an open source pipeline for cardiovascular simulation, Annals of biomedical engineering 45 (2017) 525--541.
	
	\bibitem{kikinis20133d}
	R.~Kikinis, S.~D. Pieper, K.~G. Vosburgh, 3d slicer: a platform for subject-specific image analysis, visualization, and clinical support, in: Intraoperative imaging and image-guided therapy, Springer, 2013, pp. 277--289.
	
	\bibitem{valen2018real}
	K.~Valen-Sendstad, A.~W. Bergersen, Y.~Shimogonya, L.~Goubergrits, J.~Bruening, J.~Pallares, S.~Cito, S.~Piskin, K.~Pekkan, A.~J. Geers, et~al., Real-world variability in the prediction of intracranial aneurysm wall shear stress: the 2015 international aneurysm cfd challenge, Cardiovascular engineering and technology 9 (2018) 544--564.
	
	\bibitem{deo2024few}
	Y.~Deo, F.~Lin, H.~Dou, N.~Cheng, N.~Ravikumar, A.~F. Frangi, T.~Lassila, Few-shot learning in diffusion models for generating cerebral aneurysm geometries, in: 2024 IEEE International Symposium on Biomedical Imaging (ISBI), IEEE, 2024, pp. 1--5.
	
	\bibitem{dirix2024synthesizing}
	P.~Dirix, L.~Jacobs, S.~Buoso, S.~Kozerke, Synthesizing scalable cfd-enhanced aortic 4d flow mri for assessing accuracy and precision of deep-learning image reconstruction and segmentation tasks, in: International Workshop on Simulation and Synthesis in Medical Imaging, Springer, 2024, pp. 157--166.
	
	\bibitem{saitta2023data}
	S.~Saitta, L.~Maga, C.~Armour, E.~Votta, D.~P. O’Regan, M.~Y. Salmasi, T.~Athanasiou, J.~W. Weinsaft, X.~Y. Xu, S.~Pirola, et~al., Data-driven generation of 4d velocity profiles in the aneurysmal ascending aorta, Computer Methods and Programs in Biomedicine 233 (2023) 107468.
	
	\bibitem{garzia2023coupling}
	S.~Garzia, M.~A. Scarpolini, M.~Mazzoli, K.~Capellini, A.~Monteleone, F.~Cademartiri, V.~Positano, S.~Celi, Coupling synthetic and real-world data for a deep learning-based segmentation process of 4d flow mri, Computer Methods and Programs in Biomedicine 242 (2023) 107790.
	
	\bibitem{doste2022training}
	R.~Doste, M.~Lozano, G.~Jimenez-Perez, L.~Mont, A.~Berruezo, D.~Penela, O.~Camara, R.~Sebastian, Training machine learning models with synthetic data improves the prediction of ventricular origin in outflow tract ventricular arrhythmias, Frontiers in Physiology 13 (2022) 909372.
	
	\bibitem{Myers2023MLsurrogateQSP}
	R.~C. Myers, F.~Augustin, J.~Huard, C.~M. Friedrich, Using machine learning surrogate modeling for faster qsp vp cohort generation, CPT: Pharmacometrics \& Systems Pharmacology 12~(8) (2023) 1047--1059.
	\newblock \href {https://doi.org/10.1002/psp4.12999} {\path{doi:10.1002/psp4.12999}}.
	
	\bibitem{du2022deep}
	P.~Du, X.~Zhu, J.-X. Wang, Deep learning-based surrogate model for three-dimensional patient-specific computational fluid dynamics, Physics of Fluids 34~(8) (2022) 081906.
	
	\bibitem{bruse2016non}
	J.~L. Bruse, K.~Mcleod, G.~Biglino, H.~N. Ntsinjana, C.~Capelli, T.-Y. Hsia, M.~Sermesant, X.~Pennec, A.~M. Taylor, S.~Schievano, A non-parametric statistical shape model for assessment of the surgically repaired aortic arch in coarctation of the aorta: how normal is abnormal?, in: Statistical Atlases and Computational Models of the Heart. Imaging and Modelling Challenges: 6th International Workshop, STACOM 2015, Held in Conjunction with MICCAI 2015, Munich, Germany, October 9, 2015, Revised Selected Papers 6, Springer, 2016, pp. 21--29.
	
	\bibitem{bruse2017detecting}
	J.~L. Bruse, M.~A. Zuluaga, A.~Khushnood, K.~McLeod, H.~N. Ntsinjana, T.-Y. Hsia, M.~Sermesant, X.~Pennec, A.~M. Taylor, S.~Schievano, Detecting clinically meaningful shape clusters in medical image data: metrics analysis for hierarchical clustering applied to healthy and pathological aortic arches, IEEE Transactions on Biomedical Engineering 64~(10) (2017) 2373--2383.
	
	\bibitem{scarpolini2023enabling}
	M.~A. Scarpolini, M.~Mazzoli, S.~Celi, Enabling supra-aortic vessels inclusion in statistical shape models of the aorta: a novel non-rigid registration method, Frontiers in Physiology 14 (2023) 1211461.
	
	\bibitem{cosentino2020statistical}
	F.~Cosentino, G.~M. Raffa, G.~Gentile, V.~Agnese, D.~Bellavia, M.~Pilato, S.~Pasta, Statistical shape analysis of ascending thoracic aortic aneurysm: correlation between shape and biomechanical descriptors, Journal of personalized medicine 10~(2) (2020) 28.
	
	\bibitem{sophocleous2022feasibility}
	F.~Sophocleous, A.~B{\^o}ne, A.~I. Shearn, M.~N.~V. Forte, J.~L. Bruse, M.~Caputo, G.~Biglino, Feasibility of a longitudinal statistical atlas model to study aortic growth in congenital heart disease, Computers in Biology and Medicine 144 (2022) 105326.
	
	\bibitem{Geronzi2023ShapeFeatures}
	L.~Geronzi, A.~Martinez, M.~Rochette, K.~Yan, A.~Bel-Brunon, P.~Haigron, P.~Escrig, J.~Tomasi, M.~Daniel, A.~Lalande, S.~Lin, D.~M. Marin-Castrillón, O.~Bouchot, J.~Porterie, P.~P. Valentini, M.~E. Biancolini, Computer-aided shape features extraction and regression models for predicting the ascending aortic aneurysm growth rate, Computer Biology and Medicine 162 (2023) 107052.
	\newblock \href {https://doi.org/10.1016/j.compbiomed.2023.107052} {\path{doi:10.1016/j.compbiomed.2023.107052}}.
	
	\bibitem{niederer2020creation}
	S.~A. Niederer, Y.~Aboelkassem, C.~D. Cantwell, C.~Corrado, S.~Coveney, E.~M. Cherry, T.~Delhaas, F.~H. Fenton, A.~V. Panfilov, P.~Pathmanathan, et~al., Creation and application of virtual patient cohorts of heart models, Philosophical Transactions of the Royal Society A 378~(2173) (2020) 20190558.
	
	\bibitem{mansi2011statistical}
	T.~Mansi, I.~Voigt, B.~Leonardi, X.~Pennec, S.~Durrleman, M.~Sermesant, H.~Delingette, A.~M. Taylor, Y.~Boudjemline, G.~Pongiglione, et~al., A statistical model for quantification and prediction of cardiac remodelling: Application to tetralogy of fallot, IEEE transactions on medical imaging 30~(9) (2011) 1605--1616.
	
	\bibitem{rodero2021linking}
	C.~Rodero, M.~Strocchi, M.~Marciniak, S.~Longobardi, J.~Whitaker, M.~D. O’Neill, K.~Gillette, C.~Augustin, G.~Plank, E.~J. Vigmond, et~al., Linking statistical shape models and simulated function in the healthy adult human heart, PLoS computational biology 17~(4) (2021) e1008851.
	
	\bibitem{Verstraeten2024SyntheticAVStenosis}
	S.~Verstraeten, M.~Hoeijmakers, P.~Tonino, J.~Br{\"u}ning, C.~Capelli, F.~van~de Vosse, W.~Huberts, Generation of synthetic aortic valve stenosis geometries for in silico trials, International Journal for Numerical Methods in Biomedical Engineering 40~(1) (2024) e3778.
	\newblock \href {https://doi.org/10.1002/cnm.3778} {\path{doi:10.1002/cnm.3778}}.
	
	\bibitem{oguz2016entropy}
	I.~Oguz, J.~Cates, M.~Datar, B.~Paniagua, T.~Fletcher, C.~Vachet, M.~Styner, R.~Whitaker, Entropy-based particle correspondence for shape populations, International journal of computer assisted radiology and surgery 11 (2016) 1221--1232.
	
	\bibitem{Grassi2011femurMorphing}
	L.~Grassi, N.~Hraiech, E.~Schileo, M.~Ansaloni, M.~Rochette, M.~Viceconti, Evaluation of the generality and accuracy of a new mesh morphing procedure for the human femur, Medical Engineering \& Physics 33~(1) (2011) 112--120.
	\newblock \href {https://doi.org/10.1016/j.medengphy.2010.09.014} {\path{doi:10.1016/j.medengphy.2010.09.014}}.
	
	\bibitem{Li2024ImageUncertaintyCFD}
	H.~Li, H.~Yu, J.~Wei, X.~Du, Effects of image uncertainty on blood flow simulation, in: Proceedings of the ASME 2024 International Design Engineering Technical Conferences \& Computers and Information in Engineering Conference (IDETC-CIE), 2024.
	\newblock \href {https://doi.org/10.1115/IDETC-CIE-2024-1208930} {\path{doi:10.1115/IDETC-CIE-2024-1208930}}.
	
	\bibitem{thamsen2021synthetic}
	B.~Thamsen, P.~Yevtushenko, L.~Gundelwein, A.~A. Setio, H.~Lamecker, M.~Kelm, M.~Schafstedde, T.~Heimann, T.~Kuehne, L.~Goubergrits, Synthetic database of aortic morphometry and hemodynamics: overcoming medical imaging data availability, IEEE Transactions on Medical Imaging 40~(5) (2021) 1438--1449.
	
	\bibitem{alvarez2017tracking}
	L.~Alvarez, A.~Trujillo, C.~Cuenca, E.~Gonz{\'a}lez, J.~Esclar{\'\i}n, L.~Gomez, L.~Mazorra, M.~Alem{\'a}n-Flores, P.~G. Tahoces, J.~M. Carreira, Tracking the aortic lumen geometry by optimizing the 3d orientation of its cross-sections, in: Medical Image Computing and Computer-Assisted Intervention- MICCAI 2017: 20th International Conference, Quebec City, QC, Canada, September 11-13, 2017, Proceedings, Part II 20, Springer, 2017, pp. 174--181.
	
	\bibitem{Ostendorf2024SyntheticAorticDissection}
	K.~Ostendorf, K.~B{\"a}umler, D.~Mastrodicasa, D.~Fleischmann, B.~Preim, G.~Mistelbauer, Synthetic surface mesh generation of aortic dissections using statistical shape modeling, Computers \& Graphics 124 (2024) 104070.
	\newblock \href {https://doi.org/10.1016/j.cag.2024.104070} {\path{doi:10.1016/j.cag.2024.104070}}.
	
	\bibitem{romero2021clinically}
	P.~Romero, M.~Lozano, F.~Mart{\'\i}nez-Gil, D.~Serra, R.~Sebasti{\'a}n, P.~Lamata, I.~Garc{\'\i}a-Fern{\'a}ndez, Clinically-driven virtual patient cohorts generation: an application to aorta, Frontiers in Physiology 12 (2021) 713118.
	
	\bibitem{Romero2025RobustVesselShape}
	P.~Romero, A.~Pedr{\'o}s, R.~Sebasti{\'a}n, M.~Lozano, I.~Garc{\'\i}a-Fern{\'a}ndez, A robust shape model for blood vessels analysis, Applied Mathematics and Computation 487 (2025) 129078.
	\newblock \href {https://doi.org/10.1016/j.amc.2024.129078} {\path{doi:10.1016/j.amc.2024.129078}}.
	
	\bibitem{liang2017machine}
	L.~Liang, M.~Liu, C.~Martin, J.~A. Elefteriades, W.~Sun, A machine learning approach to investigate the relationship between shape features and numerically predicted risk of ascending aortic aneurysm, Biomechanics and modeling in mechanobiology 16 (2017) 1519--1533.
	
	\bibitem{Wiputra2023thoracicAortaSSM}
	H.~Wiputra, S.~Matsumoto, J.~E. Wagenseil, A.~C. Braverman, R.~K. Voeller, V.~H. Barocas, Statistical shape representation of the thoracic aorta: accounting for major branches of the aortic arch, Computer Methods in Biomechanics and Biomedical Engineering 26~(13) (2023) 1557--1571, epub 2022 Sep 27.
	\newblock \href {https://doi.org/10.1080/10255842.2022.2128672} {\path{doi:10.1080/10255842.2022.2128672}}.
	
	\bibitem{Maquart2021BREPmeshing}
	T.~Maquart, T.~Elguedj, A.~Gravouil, M.~Rochette, 3d b‑rep meshing for real‑time data‑based geometric parametric analysis, Advanced Modeling and Simulation in Engineering Sciences 8~(1) (2021) 8.
	\newblock \href {https://doi.org/10.1186/s40323-021-00194-5} {\path{doi:10.1186/s40323-021-00194-5}}.
	
	\bibitem{young2009computational}
	A.~A. Young, A.~F. Frangi, Computational cardiac atlases: from patient to population and back, Experimental physiology 94~(5) (2009) 578--596.
	
	\bibitem{khalafvand2018assessment}
	S.~Khalafvand, J.~Voorneveld, A.~Muralidharan, F.~Gijsen, J.~Bosch, T.~van Walsum, A.~Haak, N.~de~Jong, S.~Kenjeres, Assessment of human left ventricle flow using statistical shape modelling and computational fluid dynamics, Journal of biomechanics 74 (2018) 116--125.
	
	\bibitem{metz2012regression}
	C.~T. Metz, N.~Baka, H.~Kirisli, M.~Schaap, S.~Klein, L.~A. Neefjes, N.~R. Mollet, B.~Lelieveldt, M.~de~Bruijne, W.~J. Niessen, et~al., Regression-based cardiac motion prediction from single-phase cta, IEEE transactions on medical imaging 31~(6) (2012) 1311--1325.
	
	\bibitem{goubergrits2022ct}
	L.~Goubergrits, K.~Vellguth, L.~Obermeier, A.~Schlief, L.~Tautz, J.~Bruening, H.~Lamecker, A.~Szengel, O.~Nemchyna, C.~Knosalla, et~al., Ct-based analysis of left ventricular hemodynamics using statistical shape modeling and computational fluid dynamics, Frontiers in Cardiovascular Medicine 9 (2022) 901902.
	
	\bibitem{Verhulsdonk2024ShapeOfMyHeart}
	J.~Verh{\"u}lsdonk, T.~Grandits, F.~S. Costabal, T.~Pinetz, R.~Krause, A.~Auricchio, G.~Haase, S.~Pezzuto, A.~Effland, Shape of my heart: Cardiac models through learned signed distance functions, in: Proceedings of the 7th International Conference on Medical Imaging with Deep Learning (MIDL), Vol. 250 of Proceedings of Machine Learning Research, PMLR, 2024, pp. 1584--1605.
	\newblock \href {https://doi.org/10.48550/arXiv.2308.16568} {\path{doi:10.48550/arXiv.2308.16568}}.
	
	\bibitem{Varela2017LAanalysis}
	M.~Varela, F.~Bisbal, E.~Zacur, A.~Berruezo, O.~V. Aslanidi, L.~Mont, P.~Lamata, Novel computational analysis of left atrial anatomy improves prediction of atrial fibrillation recurrence after ablation, Frontiers in Physiology 8 (2017) 68.
	\newblock \href {https://doi.org/10.3389/fphys.2017.00068} {\path{doi:10.3389/fphys.2017.00068}}.
	
	\bibitem{Bisighini2025stent}
	B.~Bisighini, M.~Aguirre, B.~Pierrat, S.~Avril, Machine learning and statistical shape modelling for real-time prediction of stent deployment in realistic anatomies, Computer Methods and Programs in Biomedicine 260 (2025) 108583.
	\newblock \href {https://doi.org/10.1016/j.cmpb.2024.108583} {\path{doi:10.1016/j.cmpb.2024.108583}}.
	
	\bibitem{hoeijmakers2020combining}
	M.~Hoeijmakers, I.~Waechter-Stehle, J.~Weese, F.~Van~de Vosse, Combining statistical shape modeling, cfd, and meta-modeling to approximate the patient-specific pressure-drop across the aortic valve in real-time, International journal for numerical methods in biomedical engineering 36~(10) (2020) e3387.
	
	\bibitem{gambaruto2012decomposition}
	A.~M. Gambaruto, D.~Taylor, D.~Doorly, Decomposition and description of the nasal cavity form, Annals of biomedical engineering 40 (2012) 1142--1159.
	
	\bibitem{keustermans2018high}
	W.~Keustermans, T.~Huysmans, F.~Danckaers, A.~Zarowski, B.~Schmelzer, J.~Sijbers, J.~J. Dirckx, High quality statistical shape modelling of the human nasal cavity and applications, Royal Society Open Science 5~(12) (2018) 181558.
	
	\bibitem{bruning2020characterization}
	J.~Br{\"u}ning, T.~Hildebrandt, W.~Heppt, N.~Schmidt, H.~Lamecker, A.~Szengel, N.~Amiridze, H.~Ramm, M.~Bindernagel, S.~Zachow, et~al., Characterization of the airflow within an average geometry of the healthy human nasal cavity, Scientific Reports 10~(1) (2020) 3755.
	
	\bibitem{Gahima2023UnfittedElasticBed}
	S.~Gahima, P.~D{\'i}ez, M.~Stefanati, J.~F. Rodr{\'\i}guez~Matas, A.~Garc{\'\i}a-Gonz{\'a}lez, \href{https://doi.org/10.3390/math11071748}{An unfitted method with elastic bed boundary conditions for the analysis of heterogeneous arterial sections}, Mathematics 11~(7) (2023) 1748.
	\newblock \href {https://doi.org/10.3390/math11071748} {\path{doi:10.3390/math11071748}}.
	\newline\urlprefix\url{https://doi.org/10.3390/math11071748}
	
	\bibitem{Ballarin2016PODGalerkinCFR}
	F.~Ballarin, E.~Faggiano, S.~Ippolito, A.~Manzoni, A.~Quarteroni, G.~Rozza, R.~Scrofani, Fast simulations of patient-specific haemodynamics of coronary artery bypass grafts based on a pod–galerkin method and a vascular shape parametrization, Journal of Computational Physics 315 (2016) 609--628.
	\newblock \href {https://doi.org/10.1016/j.jcp.2016.03.065} {\path{doi:10.1016/j.jcp.2016.03.065}}.
	
	\bibitem{vaillant2005surface}
	M.~Vaillant, J.~Glaunes, Surface matching via currents, in: Biennial international conference on information processing in medical imaging, Springer, 2005, pp. 381--392.
	
	\bibitem{lamecker2008variational}
	H.~Lamecker, Variational and statistical shape modeling for 3d geometry reconstruction, Ph.D. thesis (2008).
	
	\bibitem{liu2024survey}
	Y.~Liu, A.~Zeng, Y.~Pan, X.~Zhou, J.~Yu, A survey on 3d content generation: Nerfs, diffusion, gans, and beyond, arXiv preprint arXiv:2402.12833 (2024).
	
	\bibitem{yazdani2023survey}
	D.~Yazdani, J.~Wu, F.~Khan, L.~Wang, Y.~Zhou, G.-J. Qi, A survey on generative models: From gans to vaes and diffusion models, IEEE Transactions on Pattern Analysis and Machine Intelligence (2023).
	\newblock \href {https://doi.org/10.1109/TPAMI.2023.3290269} {\path{doi:10.1109/TPAMI.2023.3290269}}.
	
	\bibitem{Landoll2024VirtualPatientECMO}
	M.~Landoll, A.~Black, J.~Smith, E.~Johnson, Generating virtual patient data for in silico clinical trials of medical devices during extracorporeal membrane oxygenation, bioRxivPreprint posted 3 Sep 2024 (2024).
	\newblock \href {https://doi.org/10.1101/2024.09.03.610823} {\path{doi:10.1101/2024.09.03.610823}}.
	
	\bibitem{Danu2019SyntheticVesselSurfaces}
	M.~Danu, C.-I. Nita, A.~Vizitiu, C.~Suciu, L.~M. Itu, Deep learning based generation of synthetic blood vessel surfaces, in: 2019 23rd International Conference on System Theory, Control and Computing (ICSTCC), IEEE, 2019, pp. 662--667.
	
	\bibitem{Wolterink2018BloodVesselGAN}
	J.~M. Wolterink, T.~Leiner, I.~I{\v{s}}gum, \href{https://arxiv.org/abs/1804.04381}{Blood vessel geometry synthesis using generative adversarial networks}, CoRR abs/1804.04381 (2018).
	\newline\urlprefix\url{https://arxiv.org/abs/1804.04381}
	
	\bibitem{Dou2025VirtualChimera}
	H.~Dou, S.~Virtanen, N.~Ravikumar, A.~F. Frangi, A generative shape compositional framework to synthesize populations of virtual chimaeras, IEEE Transactions on Neural Networks and Learning Systems 36~(3) (2025) 4750--4764.
	\newblock \href {https://doi.org/10.1109/TNNLS.2024.3374121} {\path{doi:10.1109/TNNLS.2024.3374121}}.
	
	\bibitem{kalaie2023graphLV}
	S.~Kalaie, A.~J. Bulpitt, A.~F. Frangi, A.~Gooya, A geometric deep learning framework for generation of virtual left ventricles as graphs, in: Medical Imaging with Deep Learning (MIDL), Vol. 227 of Proceedings of Machine Learning Research, PMLR, 2023, pp. 426--443.
	\newblock \href {https://doi.org/10.48550/arXiv.2308.16568} {\path{doi:10.48550/arXiv.2308.16568}}.
	
	\bibitem{Kalaie2025RefinableShapeGeneration}
	S.~Kalaie, J.~Smith, J.~Doe, Others, An end-to-end deep learning generative framework for refinable shape matching and generation, International Journal of Computer Vision (2025).
	\newblock \href {https://doi.org/10.1007/s11263-024-02008-3} {\path{doi:10.1007/s11263-024-02008-3}}.
	
	\bibitem{Beetz2022MultiDomainVAE}
	M.~Beetz, A.~Banerjee, V.~Grau, Multi-domain variational autoencoders for combined modeling of mri-based biventricular anatomy and ecg-based cardiac electrophysiology, Frontiers in Physiology 13 (2022) 886723.
	\newblock \href {https://doi.org/10.3389/fphys.2022.886723} {\path{doi:10.3389/fphys.2022.886723}}.
	
	\bibitem{Beetz2022InterpretMeshVAE}
	M.~Beetz, J.~Corral~Acero, A.~Banerjee, I.~Eitel, E.~Zacur, T.~Lange, T.~Stiermaier, R.~Evertz, S.~J. Backhaus, H.~Thiele, A.~Bueno‑Orovio, P.~Lamata, A.~Schuster, V.~Grau, Interpretable cardiac anatomy modeling using variational mesh autoencoders, Frontiers in Cardiovascular Medicine 9 (2022) 983868.
	\newblock \href {https://doi.org/10.3389/fcvm.2022.983868} {\path{doi:10.3389/fcvm.2022.983868}}.
	
	\bibitem{Feldman2023VesselVAE}
	P.~A. Feldman, M.~Fainstein, V.~Siless, C.~Delrieux, E.~Iarussi, Vesselvae: Recursive variational autoencoders for 3d blood vessel synthesis, in: Medical Image Computing and Computer Assisted Intervention – MICCAI 2023, Vol. 14220 of Lecture Notes in Computer Science, 2023, pp. 67--76.
	\newblock \href {https://doi.org/10.1007/978-3-031-43907-0_7} {\path{doi:10.1007/978-3-031-43907-0_7}}.
	
	\bibitem{Feldman2025VesselGPT}
	P.~Feldman, M.~Sinnona, V.~Siless, C.~Delrieux, E.~Iarussi, Vesselgpt: Autoregressive modeling of vascular geometry, arXiv preprint arXiv:2505.13318 (May 2025).
	\newblock \href {https://doi.org/10.48550/arXiv.2505.13318} {\path{doi:10.48550/arXiv.2505.13318}}.
	
	\bibitem{dhariwal2021diffusion}
	P.~Dhariwal, A.~Q. Nichol, Diffusion models beat gans on image synthesis, in: Advances in Neural Information Processing Systems, Vol.~34, Curran Associates, Inc., 2021, pp. 8780--8794.
	
	\bibitem{chen2024overview}
	M.~Chen, S.~Mei, J.~Fan, M.~Wang, An overview of diffusion models: Applications, guided generation, statistical rates and optimization, arXiv preprint arXiv:2404.07771 (2024).
	
	\bibitem{du2024confild}
	P.~Du, M.~H. Parikh, X.~Fan, X.-Y. Liu, J.-X. Wang, Conditional neural field latent diffusion model for generating spatiotemporal turbulence, Nature Communications (2024).
	
	\bibitem{fan2025neural}
	X.~Fan, D.~Akhare, J.-X. Wang, Neural differentiable modeling with diffusion-based super-resolution for two-dimensional spatiotemporal turbulence, Computer Methods in Applied Mechanics and Engineering 433 (2025) 117478.
	
	\bibitem{liu2025confild}
	X.-Y. Liu, M.~H. Parikh, X.~Fan, P.~Du, Q.~Wang, Y.-F. Chen, J.-X. Wang, Confild-inlet: Synthetic turbulence inflow using generative latent diffusion models with neural fields, Physical Review Fluids 10~(5) (2025) 054901.
	
	\bibitem{Kadry2024DiffusionDigitalTwins}
	K.~Kadry, S.~Gupta, F.~R. Nezami, E.~R. Edelman, Probing the limits and capabilities of diffusion models for the anatomic editing of digital twins, npj Digital Medicine 7~(1) (2024) 354.
	\newblock \href {https://doi.org/10.1038/s41746-024-01332-0} {\path{doi:10.1038/s41746-024-01332-0}}.
	
	\bibitem{Sinha2024TrIND}
	A.~Sinha, G.~Hamarneh, Trind: Representing anatomical trees by denoising diffusion of implicit neural fields, in: Medical Image Computing and Computer‑Assisted Intervention (MICCAI), Vol. LNCS 15012, Springer Nature Switzerland, 2024, pp. 344--354.
	\newblock \href {https://doi.org/10.1007/978-3-031-72390-2_33} {\path{doi:10.1007/978-3-031-72390-2_33}}.
	
	\bibitem{Deo2024FewShotCerebralAneurysm}
	Y.~Deo, F.~Lin, H.~Dou, N.~Cheng, N.~Ravikumar, A.~F. Frangi, T.~Lassila, Few‐shot learning in diffusion models for generating cerebral aneurysm geometries, in: Proceedings of the 21st IEEE International Symposium on Biomedical Imaging (ISBI), 2024, p. 106–113.
	\newblock \href {https://doi.org/10.1109/ISBI56570.2024.10635313} {\path{doi:10.1109/ISBI56570.2024.10635313}}.
	
	\bibitem{Kuipers2024ConditionalSetDiffusion}
	T.~P. Kuipers, P.~R. Konduri, H.~A. Marquering, E.~J. Bekkers, Generating cerebral vessel trees of acute ischemic stroke patients using conditional set-diffusion, in: Proceedings of the 7th International Conference on Medical Imaging with Deep Learning (MIDL), Vol. 250 of Proceedings of Machine Learning Research, PMLR, 2024, pp. 782--792.
	
	\bibitem{Prabhakar2024VesselGraphDD}
	C.~Prabhakar, S.~Shit, F.~Musio, K.~Yang, T.~Amiranashvili, J.~C. Paetzold, H.~B. Li, B.~Menze, 3d vessel graph generation using denoising diffusion, in: Medical Image Computing and Computer-Assisted Intervention (MICCAI), Vol. 15011 of Lecture Notes in Computer Science, Springer, 2024, pp. 3--13.
	\newblock \href {https://doi.org/10.1007/978-3-031-72120-5_1} {\path{doi:10.1007/978-3-031-72120-5_1}}.
	
	\bibitem{wolterink2016dilated}
	J.~M. Wolterink, T.~Leiner, M.~A. Viergever, I.~I{\v{s}}gum, Dilated convolutional neural networks for cardiovascular mr segmentation in congenital heart disease, in: International Workshop on Reconstruction and Analysis of Moving Body Organs, Springer, 2016, pp. 95--102.
	
	\bibitem{xia2019automatic}
	Q.~Xia, Y.~Yao, Z.~Hu, A.~Hao, Automatic 3d atrial segmentation from ge-mris using volumetric fully convolutional networks, in: Statistical Atlases and Computational Models of the Heart. Atrial Segmentation and LV Quantification Challenges: 9th International Workshop, STACOM 2018, Held in Conjunction with MICCAI 2018, Granada, Spain, September 16, 2018, Revised Selected Papers 9, Springer, 2019, pp. 211--220.
	
	\bibitem{du2025ai}
	P.~Du, D.~An, C.~Wang, J.-X. Wang, Ai-powered automated model construction for patient-specific cfd simulations of aortic flows, Science Advances 11~(36) (2025) eadw2825.
	
	\bibitem{an2025hierarchical}
	D.~An, P.~Du, P.~Gu, J.-X. Wang, C.~Wang, Hierarchical log bayesian neural network for enhanced aorta segmentation, in: 2025 IEEE International Symposium on Biomedical Imaging, IEEE, 2025, pp. 1--4.
	
	\bibitem{wickramasinghe2020voxel2mesh}
	U.~Wickramasinghe, E.~Remelli, G.~Knott, P.~Fua, Voxel2mesh: 3d mesh model generation from volumetric data, in: Medical Image Computing and Computer Assisted Intervention--MICCAI 2020: 23rd International Conference, Lima, Peru, October 4--8, 2020, Proceedings, Part IV 23, Springer, 2020, pp. 299--308.
	
	\bibitem{zhao2022segmentation}
	J.~Zhao, J.~Zhao, S.~Pang, Q.~Feng, Segmentation of the true lumen of aorta dissection via morphology-constrained stepwise deep mesh regression, IEEE Transactions on Medical Imaging 41~(7) (2022) 1826--1836.
	
	\bibitem{Kong2021WholeHeartMesh}
	F.~Kong, N.~Wilson, S.~C. Shadden, A deep‐learning approach for direct whole‑heart mesh reconstruction, Medical Image Analysis 74 (2021) 102222.
	\newblock \href {https://doi.org/10.1016/j.media.2021.102222} {\path{doi:10.1016/j.media.2021.102222}}.
	
	\bibitem{sveinsson2025seqseg}
	N.~Sveinsson~Cepero, S.~C. Shadden, Seqseg: learning local segments for automatic vascular model construction, Annals of Biomedical Engineering 53~(1) (2025) 158--179.
	
	\bibitem{an2025aortadiff}
	D.~An, P.~Du, J.-X. Wang, C.~Wang, Aortadiff: Volume-guided conditional diffusion models for multi-branch aortic surface generation, arXiv preprint arXiv:2507.13404 (2025).
	
	\bibitem{zeng2022lion}
	X.~Zeng, A.~Vahdat, F.~Williams, Z.~Gojcic, O.~Litany, S.~Fidler, K.~Kreis, Lion: Latent point diffusion models for 3d shape generation, in: Advances in Neural Information Processing Systems (NeurIPS), 2022.
	
	\bibitem{Wilson2013}
	N.~M. Wilson, A.~K. Ortiz, A.~B. Johnson, The vascular model repository: A public resource of medical imaging data and blood flow simulation results, Journal of Medical Devices 7~(4) (2013) 0409231--409231.
	
	\bibitem{zhou20213d}
	L.~Zhou, Y.~Du, J.~Wu, 3d shape generation and completion through point-voxel diffusion, in: Proceedings of the IEEE/CVF international conference on computer vision, 2021, pp. 5826--5835.
	
	\bibitem{Hu2025FontanConduit}
	Z.~Hu, J.~E. Herrmann, E.~L. Schwarz, F.~M. Gerosa, N.~Emuna, J.~D. Humphrey, A.~W. Feinberg, T.-Y. Hsia, M.~A. Skylar-Scott, A.~L. Marsden, Multiphysics simulations of a bioprinted pulsatile fontan conduit, Journal of Biomechanical Engineering 147~(7) (2025) 071001.
	\newblock \href {https://doi.org/10.1115/1.4068319} {\path{doi:10.1115/1.4068319}}.
	
	\bibitem{sahni2023quantitative}
	A.~Sahni, E.~E. McIntyre, J.~D. Pal, D.~Mukherjee, Quantitative assessment of aortic hemodynamics for varying left ventricular assist device outflow graft angles and flow pulsation, Annals of Biomedical Engineering 51~(6) (2023) 1226--1243.
	
	\bibitem{antiga2008image}
	L.~Antiga, M.~Piccinelli, L.~Botti, B.~Ene-Iordache, A.~Remuzzi, D.~A. Steinman, An image-based modeling framework for patient-specific computational hemodynamics, Medical \& Biological Engineering \& Computing 46~(11) (2008) 1097--1112.
	
	\bibitem{piegl1997nurbs}
	L.~Piegl, W.~Tiller, The NURBS Book, 2nd Edition, Springer, 1997.
	\newblock \href {https://doi.org/10.1007/978-3-642-59223-2} {\path{doi:10.1007/978-3-642-59223-2}}.
	
	\bibitem{ho2020ddpm}
	J.~Ho, A.~Jain, P.~Abbeel, Denoising diffusion probabilistic models, in: Advances in Neural Information Processing Systems, Vol.~33, 2020, pp. 6840--6851.
	
	\bibitem{song2019score}
	Y.~Song, S.~Ermon, Generative modeling by estimating gradients of the data distribution, in: Advances in Neural Information Processing Systems, Vol.~32, 2019, pp. 11918--11930.
	
	\bibitem{song2021scorebased}
	Y.~Song, J.~Sohl-Dickstein, D.~P. Kingma, A.~Kumar, S.~Ermon, B.~Poole, Score-based generative modeling through stochastic differential equations, International Conference on Learning Representations (ICLR) (2021).
	
	\bibitem{ho2022classifierfree}
	J.~Ho, T.~Salimans, \href{https://arxiv.org/abs/2207.12598}{Classifier-free diffusion guidance} (2022).
	\newblock \href {http://arxiv.org/abs/2207.12598} {\path{arXiv:2207.12598}}.
	\newline\urlprefix\url{https://arxiv.org/abs/2207.12598}
	
	\bibitem{Radl2022}
	L.~Radl, Y.~Jin, A.~Pepe, J.~Li, C.~Gsaxner, F.-H. Zhao, J.~Egger, Avt: Multicenter aortic vessel tree cta dataset collection with ground truth segmentation masks, Data in Brief 40 (2022) 107801.
	
	\bibitem{OpenFOAM}
	{{Open Field Operation and Manipulation (OpenFOAM)}}, \url{http://www.openfoam.com/}.
	
\end{thebibliography}

\clearpage

\setcounter{figure}{0}
\setcounter{table}{0}
\setcounter{section}{0}
\setcounter{equation}{0}
\setcounter{algorithm}{0}
\renewcommand{\figurename}{Figure.}
\renewcommand{\thefigure}{S\arabic{figure}}
\renewcommand{\thetable}{S\arabic{table}}
\renewcommand{\thealgorithm}{S\arabic{algorithm}}
\renewcommand{\theequation}{S\arabic{equation}}
\renewcommand\thesection{Supplementary Note \arabic{section}}
\renewcommand\thesubsection{\thesection.\arabic{subsection}}
\renewcommand\thesubsubsection{\thesubsection.\arabic{subsubsection}}

\section*{Supplementary information: \\HUG-VAS: A Hierarchical NURBS-Based Generative Model for Aortic Geometry Synthesis and Controllable Editing}
\section{Details of the generative module of HUG-VAS}

\begin{figure}[htb]
    \centering
    \includegraphics[width=0.95\textwidth]{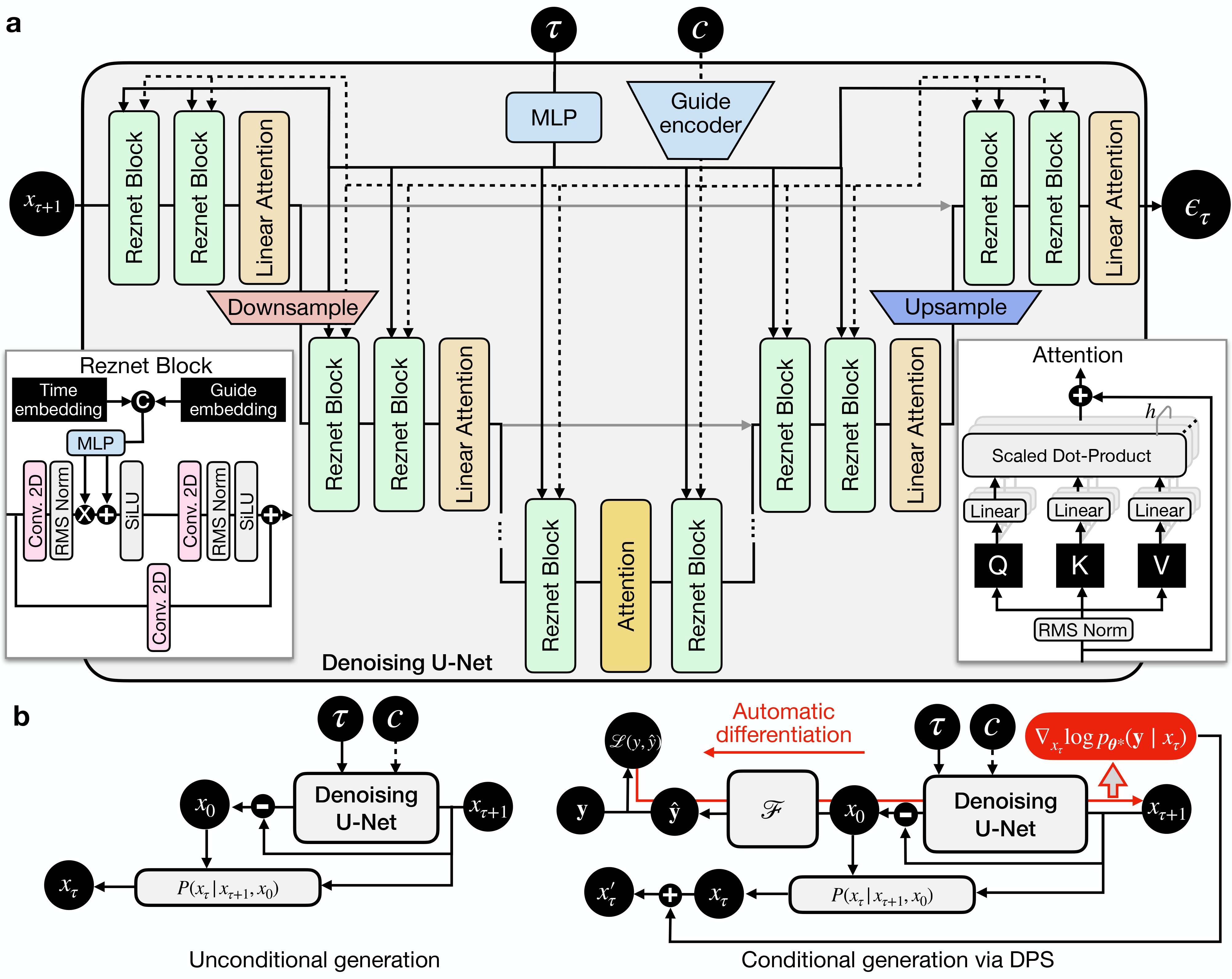}
    \caption{
    \textbf{a}, Schematic of the denoising U-Net architecture integrating time embedding, guide encoder, residual blocks, and linear attention for hierarchical feature extraction. \textbf{b}, Sampling workflows for unconditional generation (left) and conditional generation via diffusion posterior sampling (right). During conditional sampling, automatic differentiation (AD) computes the gradient of the conditional likelihood term $\nabla_{x_\tau}\log p_\theta^*(y|x_\tau)$ to guide the reverse diffusion process toward $x'_\tau$ that satisfies measurement constraints defined by the differentiable forward operator $\mathcal{F}$.}
    \label{fig:nn}
\end{figure}
The diffusion model in HUG-VAS employs a denoising U-Net that predicts the injected noise from the current noisy state \( \mathbf{x}_{\tau+1} \) (Fig.~\ref{fig:nn}a). At each scale, the network has of two ResNet blocks followed by a linear attention module. The diffusion time step $\tau$ and classifier-free conditional guidance $c$ are processed through a multilayer perceptron (MLP) and a convolutional autoencoder (encoder part only), respectively, and conditioned onto every ResNet block.

Each ResNet block contains two sub-blocks, each composed of a 2D convolutional layer, RMS normalization, and a SiLU activation, with a residual connection through an additional convolutional layer. The embedded time and classifier-free guidance features are concatenated, transformed by an MLP, and injected into the first sub-block at each scale. Note that classifier-free guidance is applied only in radius generation. Downsampling is performed using a 2D convolutional layer with a stride of 2, while upsampling is achieved through an upsampling layer followed by a 2D convolution layer with a stride of 1. Skip connections are employed between symmetric encoder–decoder scales to avoid gradient vanishing. At the bottleneck scale, a full multi-head self-attention mechanism is applied to capture self-covariance in the latent space.

The unconditional and conditional generation schemes are illustrated in Fig.~\ref{fig:nn}b. In the unconditional setting, the denoising network reconstructs the clean state \( \mathbf{x}_0 \) from \( \mathbf{x}_{\tau+1} \), which is subsequently used to compute the mean of the next denoised state \( \mathbf{x}_\tau \). In the conditional setting, a differentiable forward operator \( \mathcal{F} \) maps the clean state to the observation space, where a measurement discrepancy \( \mathcal{L}(\mathbf{y}, \hat{\mathbf{y}}) \) is evaluated. Using automatic differentiation (AD), the gradient of this loss yields a DPS-induced score correction term that guides the denoised sample toward an updated state \( \mathbf{x}'_\tau \) consistent with the target condition \( \mathbf{y} \). 

\section{Visualization of additional branches (LCCA and RCCA)}

\begin{figure}[htb]
    \centering
    \includegraphics[width=\textwidth]{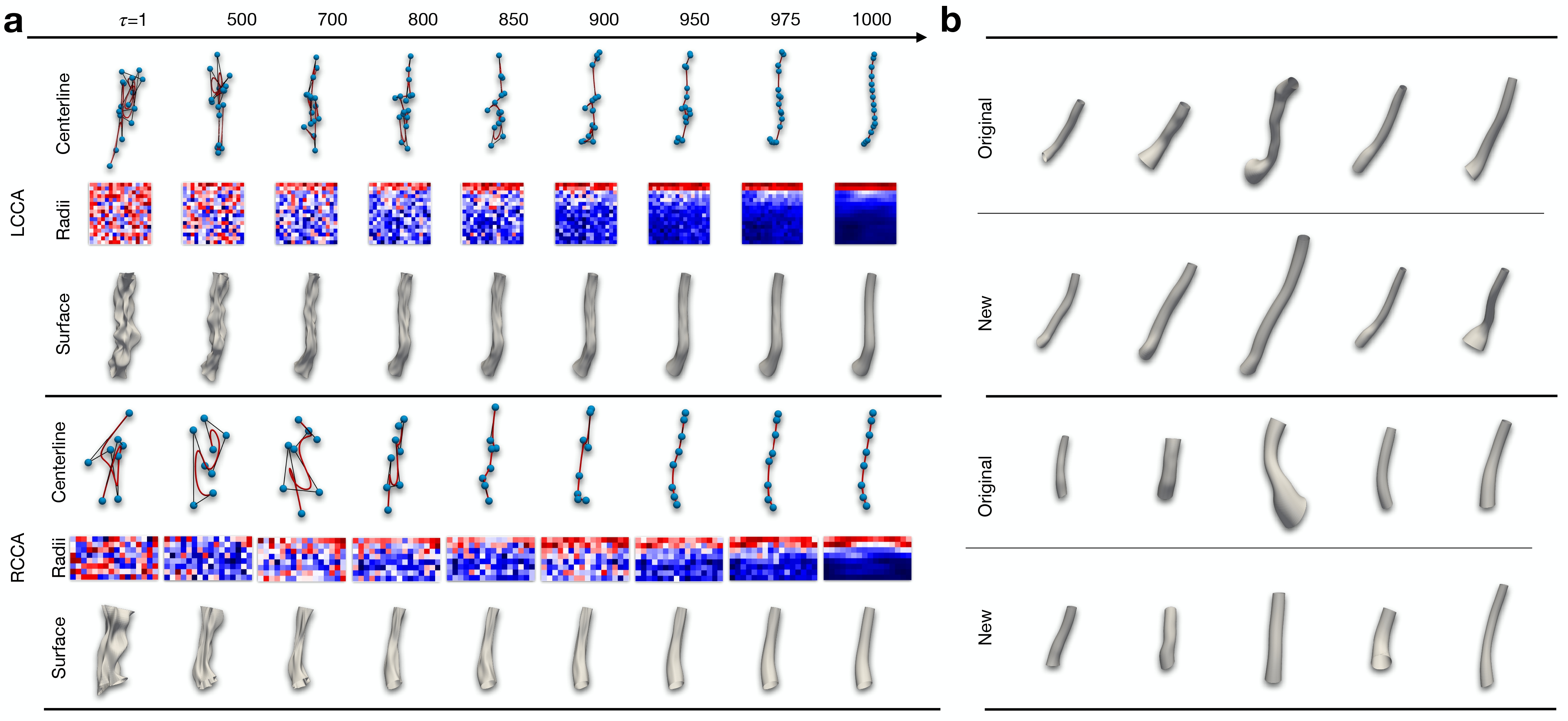}
    \caption{\textbf{a}, Visualization of the HUG-VAS denoising process for left common carotid artery (LCCA) and right common carotid artery (RCCA).\textbf{b}, Comparison between original patient-specific geometries (top rows) and newly synthesized samples
    (bottom rows) for LCCA and RCCA}
    \label{fig:RLCCA}
\end{figure}

In addition to the unconditional generation results for the aorta, LSA, and RSA, Fig.~\ref{fig:RLCCA} presents the denoising process and synthesized samples for the LCCA and RCCA branches. Panel~(a) illustrates the denoising evolution of the centerline, radius, and reconstructed surface, while Panel~(b) compares the original training samples with newly generated geometries. The denoising trajectories converge successfully to anatomically plausible centerline and radial profiles, resulting in smooth and coherent vessel surfaces for both branches. Note that we have $n=8$ controls points for RCCA due to its shorter centerline length. The generated samples exhibit notable diversity while faithfully preserving fine-grained surface features, demonstrating consistency with the original anatomical variations. 

\section{Gallery of synthesized multi-branch aortas}

\begin{figure}[htb!]
    \centering
    \includegraphics[width=0.92\textwidth]{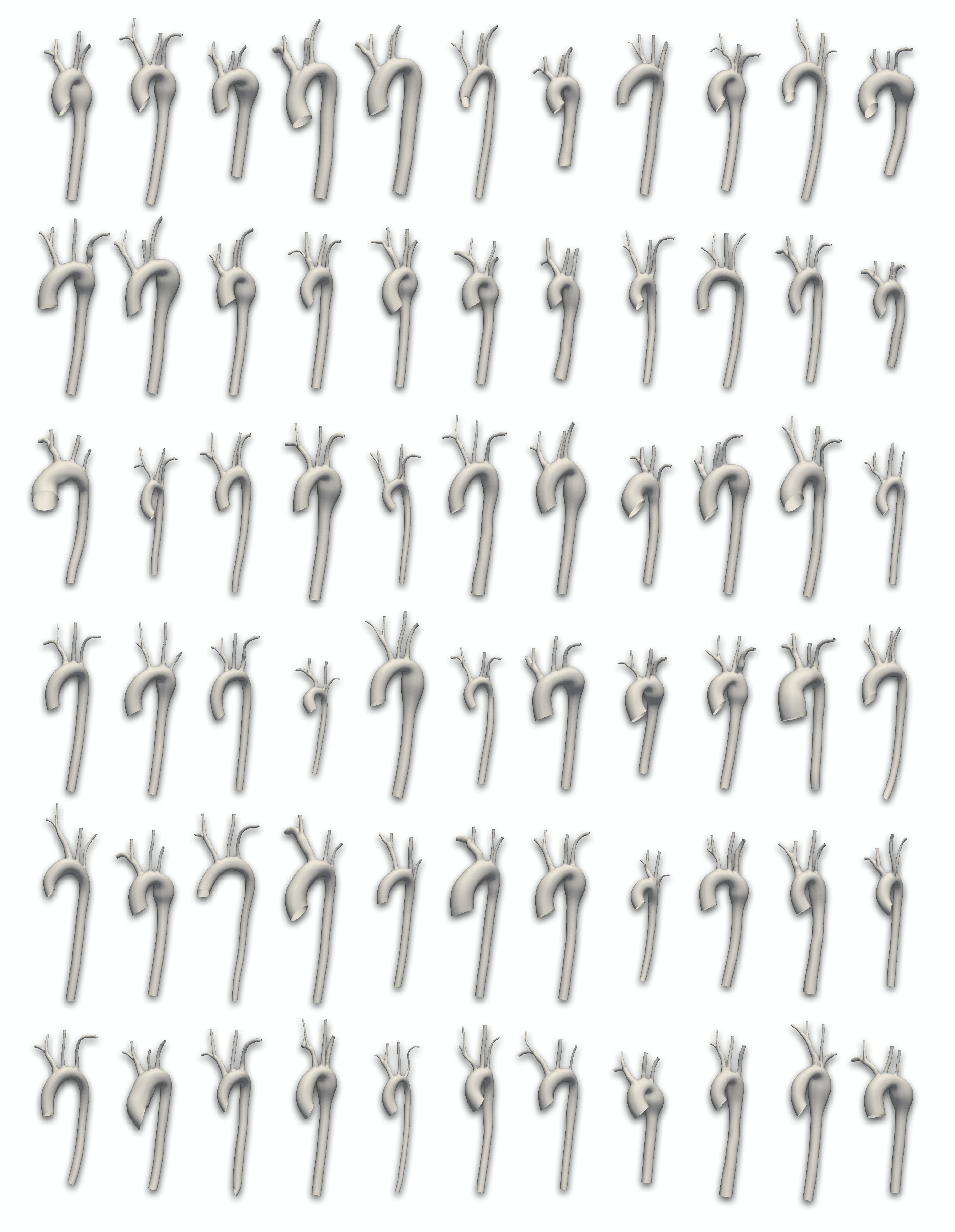}
    \caption{{Extended gallery of synthesized multi-branch aortic geometries, further supporting anatomical plausibility and morphological diversity.}}
    \label{fig:gallery}
\end{figure}

We further present 66 unconditionally generated multi-branch aorta samples in Fig.~\ref{fig:gallery}. The synthesized geometries exhibit strong anatomical plausibility and morphological diversity, while maintaining watertight surfaces suitable for downstream flow simulations. These results further demonstrate the expressivity and robustness of the proposed HUG-VAS model, underscoring its potential for data augmentation and the construction of large-scale simulated aortic flow databases. Moreover, the presence of aneurysm-like variations among the generated samples highlights its applicability for pathological analysis and risk assessment.

\newpage

\section{Generation performance against baselines}
\begin{figure}[htb]
    \centering
    \includegraphics[width=0.6\textwidth]{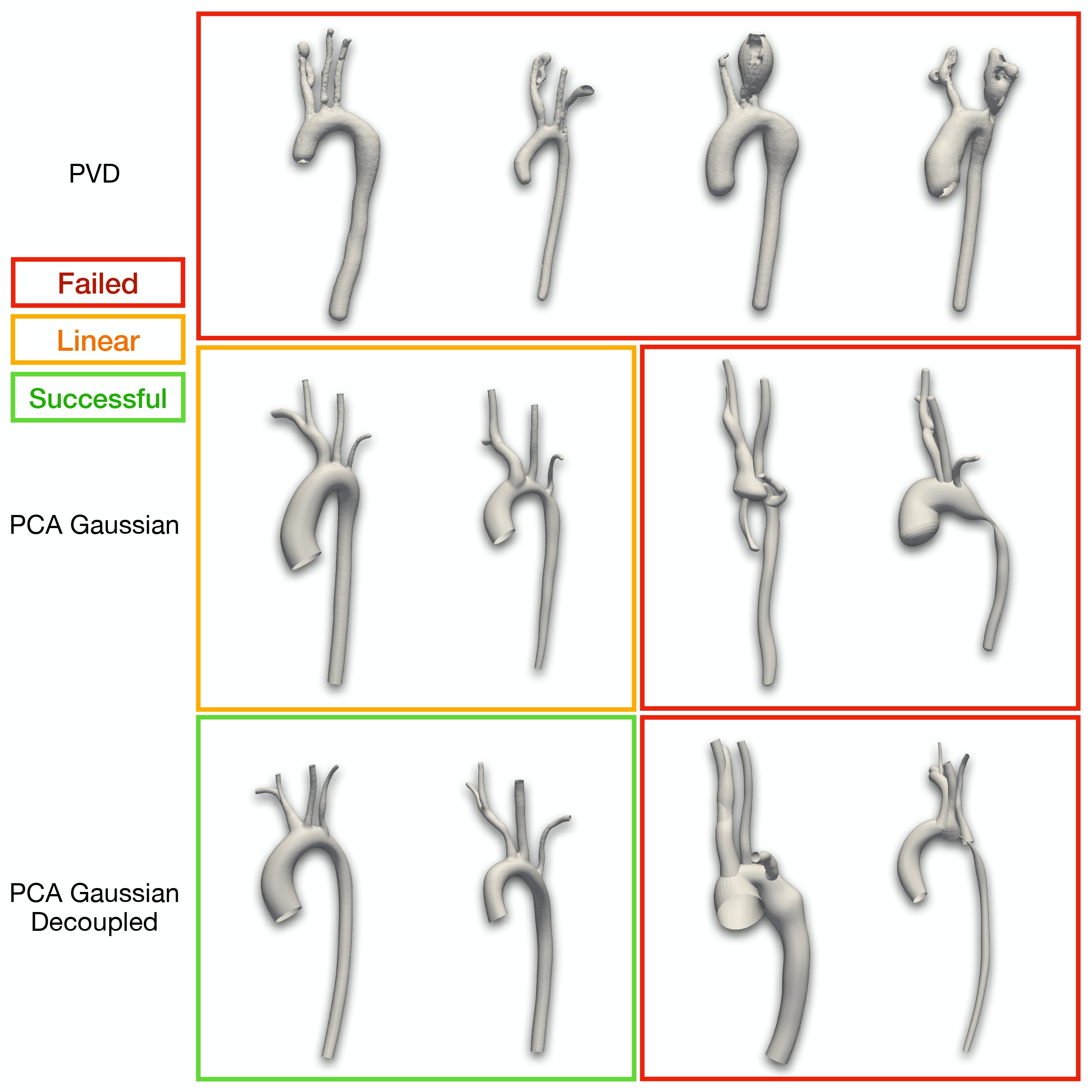}
    \caption{{Comparison of generated multi-branch aortic surfaces from the point voxel diffusion (PVD), PCA Gaussian, and PCA Gaussian–Decoupled models. Samples are classified as Failed (broken or topologically inconsistent), Similar (valid but within the linear space of training data), and Successful (novel, anatomically plausible, and watertight). The comparison highlights the superior robustness and generative diversity achieved by HUG-VAS relative to other baselines.}}
    \label{fig:baseline}
\end{figure}
Figure~\ref{fig:baseline} compares representative multi-branch aortic geometries generated by the PVD, PCA Gaussian, and PCA Gaussian–Decoupled models. Each generated surface is categorized as Failed, Linear, or Successful based on its geometric validity and novelty. Failed cases exhibit broken or non-watertight surfaces, indicating unstable or inconsistent shape generation. Linear cases produce geometrically valid shapes but remain close to the training data’s linear subspace, reflecting limited diversity. Successful cases demonstrate high anatomical plausibility and watertight topology, confirming effective shape generalization beyond the training manifold.

The PVD model, which directly denoises surface points during generation, often fails to produce intact or topologically consistent geometries. The PCA Gaussian model yields greater variability but primarily generates surfaces confined to the training linear subspace or with topological errors. The PCA Gaussian–Decoupled model can produce plausible shapes beyond the linear regime, but frequently results in severely distorted or anatomically implausible geometries due to excessive flexibility. In contrast, the proposed HUG-VAS model stably generates novel, watertight, and anatomically consistent aortic surfaces, highlighting its robustness and superior capability in capturing the intrinsic shape distribution compared with other baselines.

\section{Conditional variability of surface to centerline}
\begin{figure}[htb]
    \centering
    \includegraphics[width=\textwidth]{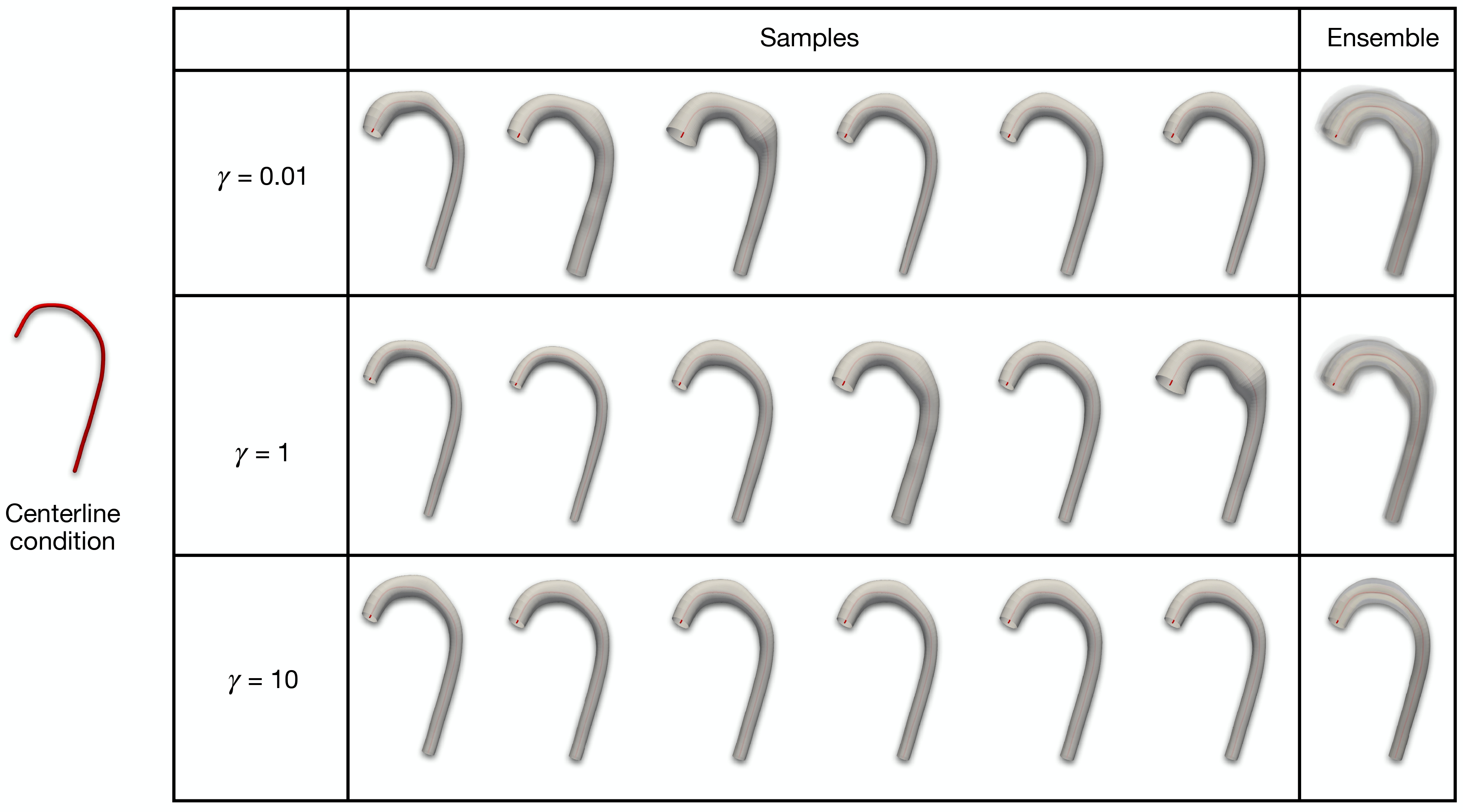}
    \caption{{Effect of conditioning strength $\gamma$ on surface generation. Surfaces are generated for a fixed centerline under varying diffusion scales ($\gamma = 0.01$, $1$, and $10$). Smaller $\gamma$ values induce greater diversity in the generated radial profiles, while larger $\gamma$ values enforce stronger centerline coupling, leading to highly consistent, low-variance geometries.}}
    \label{fig:scale}
\end{figure}

In Fig.~\ref{fig:scale}, we investigate the effect of the conditioning strength parameter $\gamma$ on surface generation for a fixed centerline. The results reveal a clear proportional relationship between $\gamma$ and the diversity of generated surfaces. A small conditioning scale ($\gamma = 0.01$) produces high variability in the radial profiles, indicating weak guidance from the centerline. As $\gamma$ increases, the generated surfaces become more consistent and tightly constrained by the centerline, with $\gamma = 10$ yielding minimal uncertainty and a nearly deterministic centerline–radius relationship. This tunable conditional variability arises from the hierarchical design of the vessel parameterization, a feature unique to our framework and not achievable with previous SSM methods.

\section{Semi-automatic segmentation software}
\begin{figure}[htb]
    \centering
    \includegraphics[width=0.8\textwidth]{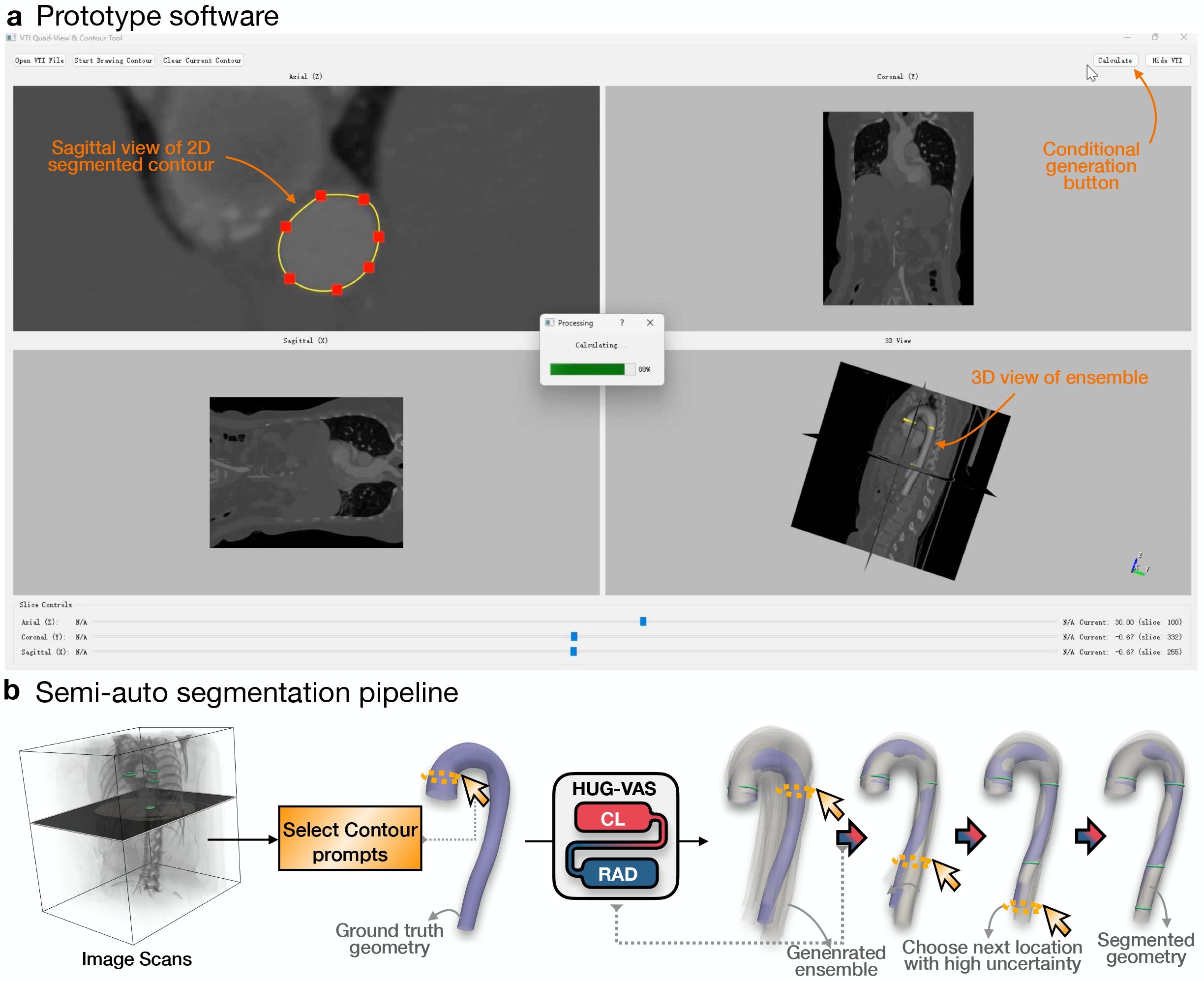}
    \caption{{\textbf{a}, Interface of the semi-automatic segmentation software showing axial, coronal, and sagittal 2D views alongside a 3D rendering of the generated ensemble. Users can draw contour prompts and trigger conditional generation via the “Calculate” button for real-time surface synthesis. \textbf{b}, Illustration of the semi-automatic segmentation workflow. The user iteratively selects contour prompts at regions of high uncertainty, progressively refining the generated ensemble until convergence to the ground-truth surface. The proposed pipeline achieves over fourfold speed-up compared with the standard SimVascular segmentation workflow.}}
    \label{fig:UI}
\end{figure}

To demonstrate the proposed interactive semi-automatic segmentation pipeline, we developed a custom segmentation software with a graphical user interface. A screenshot of the interface is shown in Fig.~\ref{fig:UI}a, displaying the axial (top left), coronal (top right), and sagittal (bottom left) 2D views, along with a 3D rendering (bottom right) of the image data. Users can freely adjust slice positions using the sliders at the bottom of the window. Manual contour prompts can be drawn directly on any 2D view, after which the “Calculate” button triggers conditional generation based on the selected contours. The generated surface ensemble is immediately rendered across all views for real-time visualization.

The recommended semi-automatic segmentation workflow is illustrated in Fig.~\ref{fig:UI}b. Starting from left to right, the user first segment one contour from axial image slices and use it as the guiding prompt to perform conditional generation. The resulting ensemble (second vessel plot) reveals spatially varying uncertainty across the surface. The user then provides an additional contour at a region with high uncertainty, leading to a refined ensemble (third vessel plot). By iteratively supplying contours at locations of maximal uncertainty, the ensemble progressively concentrates toward the ground-truth surface (purple). In practice, only four contour prompts are sufficient to suppress surface uncertainty and achieve an accurate segmentation. This process is substantially faster than the traditional workflow in SimVascular where much more cross-sectional contours need to be provided. 

A full demonstration video of this segmentation process is provided in Supplementary Video 1, and a side-by-side comparison with the SimVascular workflow is shown in Supplementary Video 2, where HUG-VAS achieves more than a fourfold speed-up. The conditional generation process can be further accelerated through advanced sampling strategies, offering potential for even greater efficiency in the proposed semi-automatic segmentation framework.

\section*{Supplementary Videos}

\textbf{Supplementary Video 1.}
Demonstration of the semi-automatic vascular segmentation workflow using HUG-VAS. 
The video shows interactive contour creation on 2D image slices, which is used as a conditional prompt for HUG-VAS to perform real-time conditional generation of a 3D vascular surface ensemble.

\textbf{Supplementary Video 2.}
Side-by-side comparison of vascular segmentation workflows using HUG-VAS and SimVascular on the same image data, demonstrating the substantial reduction in user interaction and overall segmentation time achieved by HUG-VAS.

\clearpage
\end{document}